\documentclass[iicol,pdflatex,sn-basic]{sn-jnl}% Default with double column layout

\jyear{2024}%

\theoremstyle{thmstyleone}%
\theoremstyle{thmstyletwo}%

\theoremstyle{thmstylethree}%

\usepackage{graphicx}
\usepackage{amsmath}
\usepackage{amssymb}
\usepackage{booktabs}

\usepackage{amsfonts}
\usepackage{bbding}
\usepackage{caption}
\usepackage{subcaption}

\usepackage{colortbl}
\usepackage{multirow}
\usepackage{xspace}
\usepackage{xr}

\usepackage{pifont}% http://ctan.org/pkg/pifont

\usepackage{comment}
\usepackage{xcolor}
\usepackage[capitalize]{cleveref}
\crefname{section}{Sec.}{Secs.}
\Crefname{section}{Section}{Sections}
\Crefname{table}{Table}{Tables}
\crefname{table}{Table}{Tables}

\newcommand{\etal}{\textit{et al.} }

\usepackage{ulem}

\begin{document}
%%=============================================================%%

\setcounter{page}{0}
\setcounter{section}{0}
\setcounter{table}{0}
\setcounter{figure}{0}

\newpage
\title[Rethinking Open-Set Object Detection]{\color{black} Rethinking Open-Set Object Detection: Issues, a New Formulation, and Taxonomy}


\author*[1]{\fnm{Yusuke} \sur{Hosoya}}\email{yhosoya@vision.is.tohoku.ac.jp}
\author[1]{\fnm{Masanori} \sur{Suganuma}}\email{suganuma@vision.is.tohoku.ac.jp}
\author[1,2]{\fnm{Takayuki} \sur{Okatani}}\email{okatani@vision.is.tohoku.ac.jp}


% \affil*[1]{\orgdiv{Department}, \orgname{Organization}, \orgaddress{\street{Street}, \city{City}, \postcode{100190}, \state{State}, \country{Country}}}
\affil[1]{\orgdiv{GSIS}, \orgname{Tohoku University}, \orgaddress{\city{Sendai}, \postcode{980-8579}, \state{Miyagi}, \country{Japan}}}
\affil[2]{\orgname{RIKEN Center for AIP}, \orgaddress{\city{Tokyo}, \postcode{103-0027}, \country{Japan}}}
% \affil[3]{\orgdiv{Department}, \orgname{Organization}, \orgaddress{\street{Street}, \city{City}, \postcode{610101}, \state{State}, \country{Country}}}

\abstract{
Open-set object detection (OSOD), a task involving the detection of unknown objects while accurately detecting known objects, has recently gained attention. However, we identify a fundamental issue with the problem formulation employed in current OSOD studies.
Inherent to object detection is knowing ``what to detect,'' which contradicts the idea of identifying ``unknown'' objects. 
This sets OSOD apart from open-set recognition (OSR).
This contradiction complicates a proper evaluation of methods' performance, a fact that previous studies have overlooked. 
Next, we propose a novel formulation wherein detectors are required to detect both known and unknown classes within specified super-classes of object classes. This new formulation is free from the aforementioned issues and has practical applications. Finally, we design benchmark tests utilizing existing datasets and report the experimental evaluation of existing OSOD methods. 
The results show that existing methods fail to accurately detect unknown objects due to misclassification of known and unknown classes rather than incorrect bounding box prediction.
As a byproduct, we introduce a taxonomy of OSOD, resolving confusion prevalent in the literature. We anticipate that our study will encourage the research community to reconsider OSOD and facilitate progress in the right direction. 
}

\maketitle

%===========================================================
\section{Introduction}\label{sec:introduction}

Open-set object detection (OSOD)\footnote{Recent studies (e.g., \citep{GroundingDINO}) use the term OSOD interchangeably with ``open-vocabulary object detection''. In this paper, we adhere to the traditional definition of OSOD \citep{Overlooked,Opendet}, strictly in the sense defined above. For more on the differences/relationships between the two, see Sec~\ref{sec:ovd}.} is the problem of correctly detecting known objects in images while adequately dealing with unknown objects (e.g., detecting them as unknown). Here, known objects are the class of objects that detectors have seen at training time, and unknown objects are those they have not seen before. It has attracted much attention recently \citep{DropoutSample,Overlooked,ORE,OW-DETR,ORDER,Opendet,OSOD_W}. 

% -------------------------------------------------
\begin{figure*}[t]
\centering
\includegraphics[width=0.98\linewidth]{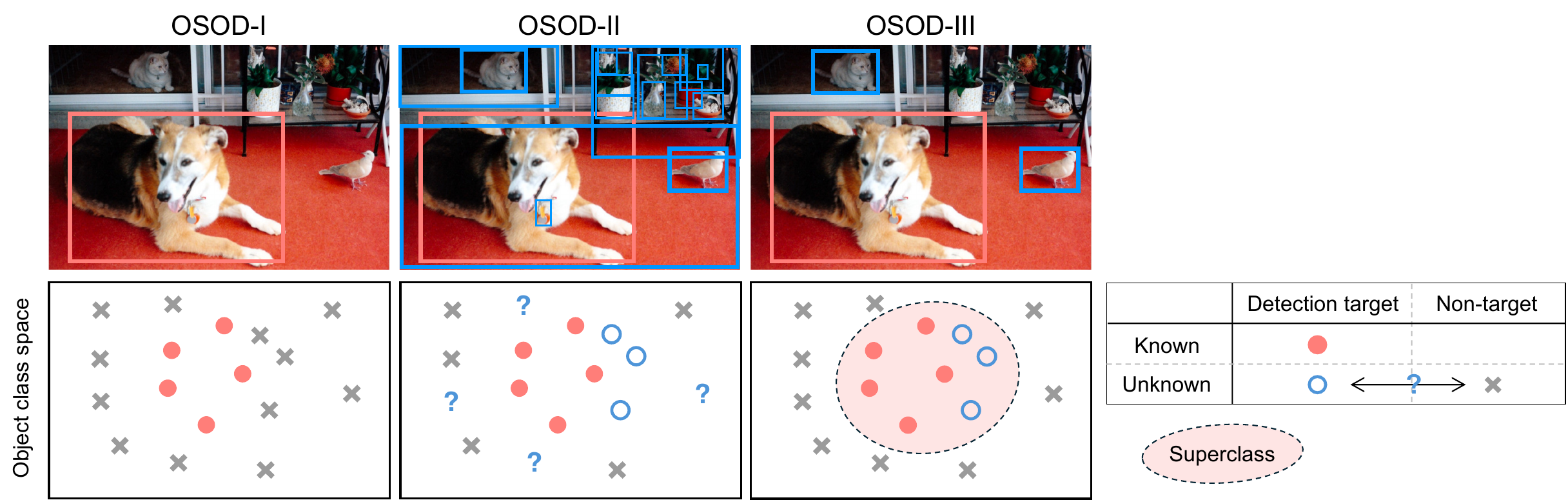}
\vspace*{2mm}
\caption{Illustration of OSOD-I, -II, and -III with image examples (top) and object class space (bottom). OSOD-I: Detect the known objects without being distracted by unknown objects. OSOD-II: Detect known and unknown objects as such, although `objectness'---what should or should not be a detection target---is ambiguous unless explicitly defined. OSOD-III: Detect known and unknown objects belonging to the same super-class as such.}
\label{fig:OSOD_toxm}
\end{figure*}
% -------------------------------------------------

% -------------------------------------------------
\renewcommand{\arraystretch}{1.5}
\begin{table*}[t]
\footnotesize\centering
\caption{Proposed categorization of OSOD problems. ``Det. target'' indicates the target of detection. K and U indicate known and unknown objects, respectively.}
\label{tbl:OSOD_toxm}
\scalebox{1.1}{
    \begin{tabular}{c|c|c|c}\specialrule{0.8pt}{0pt}{0pt}\hline  
    \multicolumn{1}{c|}{Type} & \multicolumn{1}{c|}{Det. target} & \multicolumn{1}{c|}{Unknown} & \multicolumn{1}{c}{Evaluation} \\ \hline\hline
    OSOD-I \citep{DropoutSample,Overlooked} & K & Any classes & Feasible \\
    OSOD-II \citep{ORE,Opendet} & K+U & Any classes & Hard \\ 
    \rowcolor{gray!30}{\textbf{OSOD-III}} & K+U & 
    \begin{tabular}{c}
    Any sub-classes in \\ a known super-class
    \end{tabular} & Feasible \\ \specialrule{0.8pt}{0pt}{0pt}\hline
    \end{tabular}
}
\end{table*}
\renewcommand{\arraystretch}{1}
% -------------------------------------------------

Early studies of OSOD \citep{DropoutSample,Uncertain_ObjDet,Overlooked} consider how accurately detectors can detect known objects, without being distracted by unknown objects present in input images, which we will refer to as OSOD-I in what follows. Recent studies \citep{ORE,OW-DETR,ORDER,Opendet,OSOD_W} have shifted the focus to detecting unknown objects as well. They follow the studies of open-set recognition (OSR) \citep{org_OSR,OpenMax,C2AE,ClosedSet_ICLR,PROSER} and aim to detect any arbitrary unknown objects while preserving detection accuracy for known-classes, which we will refer to as OSOD-II; see Fig.~\ref{fig:OSOD_toxm}.

In this paper, we first point out a fundamental issue with the problem formulation of OSOD, which many recent studies rely on, specifically OSOD-II as defined above. OSOD-II requires detectors to detect both known-class and unknown-class objects. However, since unknown-class objects belong to an open set and can encompass any arbitrary classes, it is impossible for detectors to be fully aware of what to detect and what not to detect during inference. To address this, a potential approach is to design a detector that detects any ``objects'' appearing in images and classifies them as either known or unknown classes. However, this approach is not feasible due to the ambiguity in the definition of ``objects.'' For instance, should the tires of a car be considered as objects? 
It is important to note that such a difficulty does not arise in OSR since it is classification; 
it involves classifying a single input image as either known or unknown. In this setting, anything that is not known is automatically defined as unknown---even if it technically belongs to an open set.
Additionally, the aforementioned issue makes it hard to evaluate the performance of methods. Existing studies employ metrics such as A-OSE \citep{DropoutSample} and WI \citep{Overlooked}, which primarily measure the accuracy of {\em known} object detection (i.e., OSOD-I) and are not suitable for evaluating unknown object detection with OSOD-II.

In light of the above, this paper next introduces a new problem formulation for OSOD, named OSOD-III. This formulation, previously neglected in existing research, is of substantial practical importance. 
OSOD-III uniquely focuses on unknown classes that are part of the same super-classes as the known classes, differentiating it from OSOD-II (as detailed in Table \ref{tbl:OSOD_toxm} and Fig.~\ref{fig:OSOD_toxm}). 
An illustrative application is a traffic sign detector used in advanced driver-assistance systems (ADAS), which, having been pre-trained on existing traffic signs, is tasked with identifying newly introduced traffic signs as novel entities.

We design benchmark tests for OSOD-III using three existing datasets: Open Images \citep{OpenImages}, Caltech-UCSD Birds-200-2011 (CUB200) \citep{CUB200}, and Mapillary Traffic Sign Dataset (MTSD) \citep{MTSD}. 
We then evaluate the performance of five recent methods (designed for OSOD-II), namely ORE \citep{ORE}, Dropout Sampling (DS) \citep{DropoutSample}, VOS \citep{VOS}, OpenDet \citep{Opendet}, {\color{black} and OrthogonalDet \citep{OrthogonalDet}}.  We also test a naive baseline method that classifies predicted boxes as known or unknown based on a simple uncertainty measure computed from predicted class scores. The results yield valuable insights. Firstly, the previous methods known for their good performance in metrics such as A-OSE and WI performed similarly or even worse than our simple baseline when they are evaluated with average precision (AP) in unknown object detection, a more appropriate performance metric. It is worth mentioning that our baseline employs standard detectors trained conventionally, without any additional training steps or extra architectures. Secondly, and more importantly, additional improvements are necessary to enable practical applications of OSOD(-III).

Our contributions are summarized as follows:
\begin{itemize}
    \item We highlight a fundamental issue with the problem formulation used in current OSOD studies, which renders it ill-posed and makes proper performance evaluation difficult. 
    \item We introduce a new variation of OSOD, named OSOD-III, which 
    has been overlooked in the literature, yet  holds practical importance.
    \item We develop benchmark tests for OSOD-III using existing public datasets and present detailed analyses on the performance of existing OSOD methods.
\end{itemize}

%===========================================================
\section{Rethinking Open-set Object Detection}\label{sec:rethingkingOSOD}

\subsection{Formalizing Problems}\label{sec:formalizing_problems}

We first formulate the problem of open-set object detection (OSOD). Previous studies refer to two different problems as OSOD without clarification. We use the names of OSOD-I and -II to distinguish the two, which are defined as follows.

\smallskip
\noindent
{\bf OSOD-I}~~
{\em The goal is to detect all instances of known objects in an image without being distracted by unknown objects present in the image. We want to avoid mistakenly detecting unknown object instances as known objects.}  

\smallskip
\noindent
{\bf OSOD-II}~
{\em The goal is to detect all instances of known and unknown objects in an image, identifying them correctly (i.e., classifying them to known classes if known and to the ``unknown'' class otherwise).} 

\medskip
OSOD-I and -II both consider applying a closed-set object detector (i.e., a detector trained on a closed-set of object classes) to an open-set environment where the detector encounters objects of unknown class.
Their difference is whether or not the detector detects unknown objects. OSOD-I does not; its concern is with the accuracy of detecting known objects. This problem is first studied in \citep{Overlooked,Uncertain_ObjDet,DropoutSample}. 
On the other hand, OSOD-II detector detects unknown objects as well, and thus their detection accuracy matters. OSOD-II is often considered as a part of open-world object detection (OWOD)
\citep{ORE,OW-DETR,RevisitOWOD,ORDER,UC-OWOD}.

The existing studies of OSOD-II rely on OWOD \citep{ORE} for the problem formulation, which aims to generalize the concept of OSR (open-set recognition) to object detection. In OSR, {\em unknown} means ``anything but known''. Its direct translation to object detection is that {\em any arbitrary classes of objects but known objects can be considered unknown.} 
This formulation is reflected in the experimental settings employed as a common benchmark test in these studies. 
Table \ref{tbl:cat_split_owod} shows the setting, which treats the 20 object classes of PASCAL VOC \citep{VOC} as known classes and non-overlapping 60 classes from 80 of COCO \citep{MSCOCO} as unknown classes.
For instance, the first split comprises the 20 classes from the PASCAL VOC dataset \citep{VOC}, the second split encompasses classes related to outdoor items, accessories, and home appliances, while the third split consists of classes pertaining to sports and food.
This division of classes underscores the basic assumption that known and unknown objects are largely unrelated.

% -------------------------------------------------
\renewcommand{\arraystretch}{1.5}
\begin{table*}[t]
\centering
\caption{The class split employed in the standard benchmark test employed in recent studies of OSOD \citep{ORE,OW-DETR,Opendet,RevisitOWOD,OSOD_W,UC-OWOD}. The numbers in parentheses indicate the number of categories.
}
\label{tbl:cat_split_owod}
\scalebox{0.8}{
    \begin{tabular}{c|cccc} \specialrule{0.7pt}{0pt}{0pt}\hline
        & Split1 & Split2 & Split3 & Split4 \\ \hline
        \begin{tabular}{c}
            Classes
        \end{tabular} & 
        \begin{tabular}{c}
            PASCAL VOC\\
            objects (20)
        \end{tabular} & 
        \begin{tabular}{c}
            Outdoor(5), Accessories(5), \\
            Appliance(5), {\color{black} Animal(4)}, {\em Truck}
        \end{tabular} &
        \begin{tabular}{c}
            Sports(10), Food(10)
        \end{tabular} & 
        \begin{tabular}{c}
            Electronic(5), Indoor(7), \\
            Kitchen(6), Furniture(2)
        \end{tabular} \\ \specialrule{0.7pt}{0pt}{0pt}\hline
    \end{tabular}
}
\end{table*}
\renewcommand{\arraystretch}{1}
% -------------------------------------------------

However, this OSOD-II's formulation has an issue, making it ill-posed. It is because the task is detection. Detectors are requested to detect only objects that should be detected. It is a primary problem of object detection to judge whether or not something should be detected. What should not be detected include objects belonging to the background and irrelevant classes. Detectors learn to make this judgment, which is feasible for a closed set of object classes; what to detect is specified. However, this does not apply to OSOD-II, which aims at detecting also unknown objects defined as above. It is infeasible to specify what to detect and what not for any arbitrary objects in advance.

\begin{figure}[t]
\centering
\includegraphics[width=1.0\linewidth]{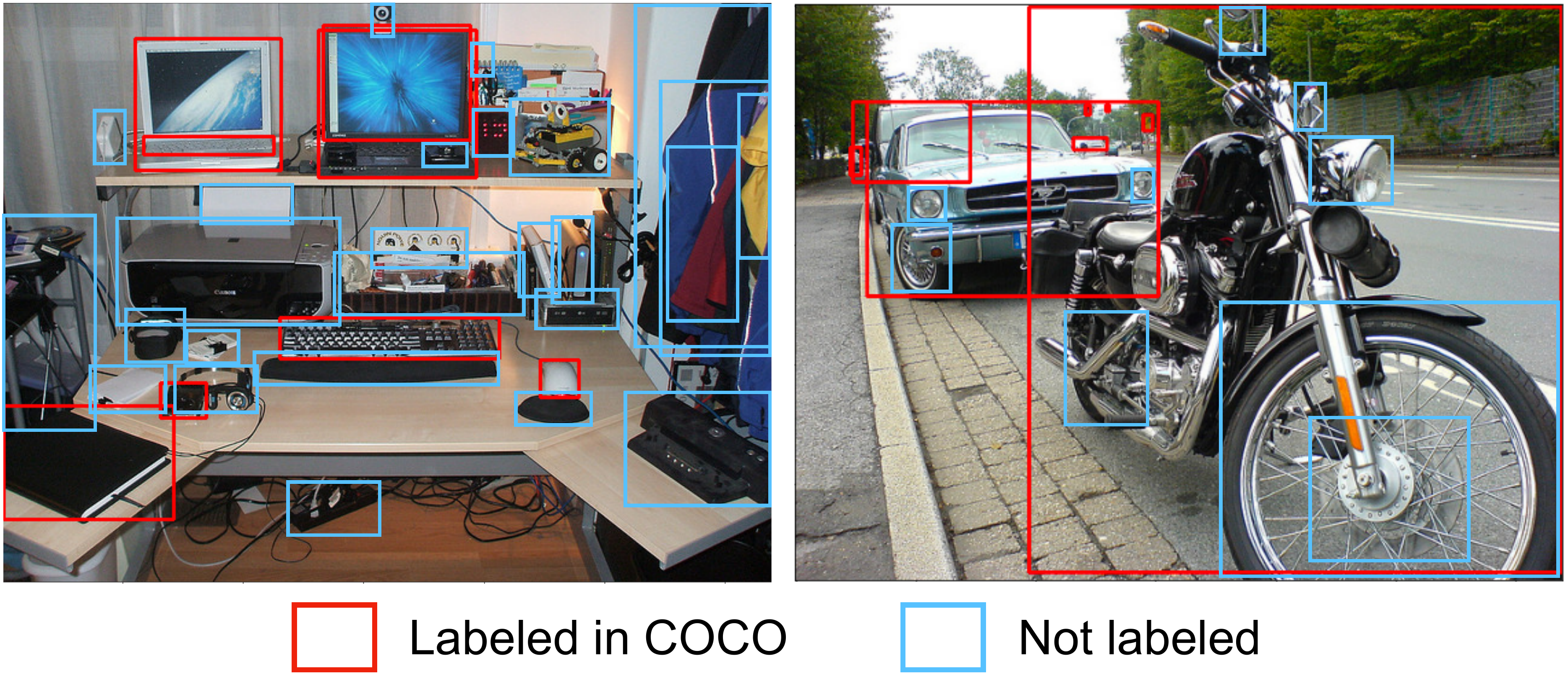}
\caption{
Example images showing that ``object'' is an ambiguous concept. It is impractical to cover an unlimited range of object instances with a finite set of predefined categories.
}
\label{fig:ambg_obj}
\end{figure}

A naive solution to this difficulty is to detect {\em any} objects as long as they are ``objects.'' However, it is not practical since defining what an object is itself hard.
Figure~\ref{fig:ambg_obj} provides examples from COCO images. COCO covers only 80 object classes (shown in red rectangles in the images), and many unannotated objects are in the images (shown in blue rectangles). Is it necessary to consider every one of them?
Moreover, it is sometimes subjective to determine what constitutes individual ``objects.''
For instance, a car consists of multiple parts, such as wheels, side mirrors, and headlights, which we may want to treat as ``objects’’ depending on applications. 
This difficulty is well recognized in the prior studies of open-world detection \citep{ORE,OW-DETR} and zero-shot detection \citep{ZeroShotObjDet,VRD_LP}.

% -------------------------------------------------
% [for two columns]
% -------------------------------------------------
% -------------------------------------------------
% [for two columns]
\begin{figure}[t]
\centering
\includegraphics[width=1.0\linewidth]{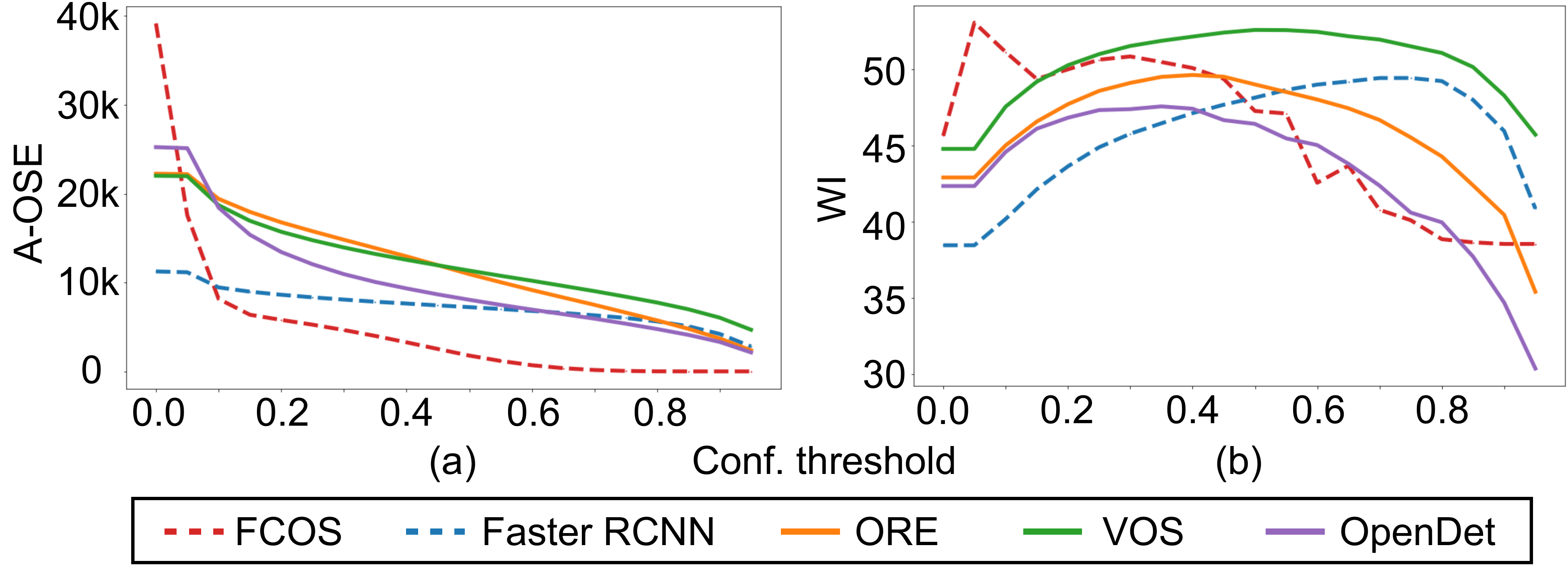}
\caption{
A-OSE (a) and WI (b) of different methods at different detector operating points. Smaller values mean better performance for both metrics. The horizontal axis indicates the confidence threshold for selecting bounding box candidates. Methods' ranking varies on the choice of the threshold.
}
\label{fig:aose_wi_plot}
\end{figure}
% -------------------------------------------------

\subsection{Metrics for Measuring OSOD Performance}\label{sec:wi_aose_defacts}

The above difficulty also leads to make it hard to evaluate how well detectors detect unknown objects. The previous studies of OSOD employ two metrics for evaluating methods' performance, i.e., absolute open-set error (A-OSE) \citep{DropoutSample} and wilderness impact (WI) \citep{Overlooked}. A-OSE is the number of predicted boxes that are in reality unknown objects but wrongly classified as known classes \citep{DropoutSample}. WI measures the ratio of the number of erroneous detections of unknowns as knowns (i.e., A-OSE) to the total number of detections of known instances, given by
\begin{equation}
\label{eq:wi}
  {\rm WI} = \frac{P_{K}}{P_{K \cup U}} - 1  = \frac{\rm{A\text{-}OSE}}{\mathrm{TP}_\mathit{known}+\mathrm{FP}_\mathit{known}},
\end{equation}
where $P_{K}$ indicates the precision measured in the close-set setting; $P_{K \cup U}$ is that measured in the open-set setting; and
$\mathrm{TP}_\mathit{known}$ and $\mathrm{FP}_\mathit{known}$ are the number of true positives and false positives  for known classes, respectively.

These two metrics are originally designed for OSOD-I; they evaluate detectors' performance in open-set environments. Precisely, they measure how frequently a detector wrongly detects and misclassifies unknown objects as known classes (lower is better).

Nevertheless, previous studies of OSOD-II have employed A-OSE and WI as primary performance metrics. We point out that these metrics are pretty insufficient to evaluate OSOD-II detectors since they cannot evaluate the accuracy of detecting unknown objects, as mentioned above.
They evaluate only one type of error, i.e., detecting unknown as known, and ignore the other type of error, detecting known as unknown. 

In addition, we point out that A-OSE and WI are not flawless even as OSOD-I performance metrics. That is, they merely measure the detectors' performance at a single operating point; they cannot take the precision-recall tradeoff into account, the fundamental nature of detection. Specifically, previous studies \citep{ORE} report A-OSE values for bounding boxes with confidence score $\geq 0.05$\footnote{This is not clearly stated in the literature but can be confirmed with the public source code in GitHub repositories, e.g., \url{https://github.com/JosephKJ/OWOD}.}. As for WI, previous studies \citep{ORE,OW-DETR,Opendet,ORDER} choose the operating point of recall $= 0.8$. 
Thus, they show performance only partially since the setting is left to end users. 
Figures \ref{fig:aose_wi_plot}(a) and (b) show the profiles of A-OSE and WI, respectively, over the possible operating points of several existing OSOD-II detectors. It is seen that the ranking of the methods varies depending on the choice of confidence threshold. 

In summary, A-OSE and WI are insufficient for evaluating OSOD-II performance since i) they merely measure OSOD-I performance, i.e., only one of the two error types, and ii) they are metrics at a single operating point. To precisely measure OSOD-II performance, we must use average precision (AP), the standard metric for object detection, also to evaluate unknown object detection. It should be noted that while all the previous studies of OSOD-II report APs for known object detection, only a few report APs for unknown detection, such as \citep{Opendet,UC-OWOD}, probably because of the mentioned difficulty of specifying what unknown objects to detect and what not. 

%===========================================================
\subsection{Relation to Open-vocabulary Detection}\label{sec:ovd}

Open-vocabulary object detection (OVD) \citep{OVD_Cap,OVD_Distill} is a problem of detecting specific object classes in a zero-shot manner by using text to indicate their class names and/or attributes. This problem has recently gained significant attention \citep{detic,GroundingDINO,CORA,GLIP,simple_ovd,detclip,detclipv2}. Since the purpose of OVD can be said to eliminate the concept of unknown classes, some might think that if OVD is fully developed, the term ``unknown'' would become obsolete, questioning the relevance of OSOD. However, this is not accurate for several reasons.

First, not all objects targeted for detection can be effectively described in words. For example, the diverse types of damages in factory-produced products are often indescribable in language, sometimes only identifiable as `Damage Type No.1, No.2,' etc. This limitation extends to newly created traffic signs, company logos, and more. Furthermore, since these items are constantly emerging worldwide, the database of object classes in OVD, which relies heavily on internet-sourced corpora (such as those used in CLIP \citep{CLIP}), can become outdated quickly. There are also concerns about the efficiency of OVD, particularly when implemented on resource-constrained hardware, like in automotive applications.

Conversely, the fundamental elements of both OVD and OSOD may reside in the structure of their feature spaces. This commonality suggests that integrating the two methods could potentially yield mutually beneficial solutions. However, further investigation into this prospect is a subject for future research.

\subsection{Summary of Issues with OSOD-II}\label{sec:osod2_summary}

Our central claim in this section is:
{\em OSOD-II cannot be reliably performed unless the notion of ``object'' is defined explicitly.}
We add three important clarifications:
\begin{itemize}
    \item Objectness is often intrinsically ambiguous and therefore hard to specify precisely.
    \item The benchmarks on which most OSOD-II studies rely (e.g., COCO-based datasets) do not resolve this ambiguity.
    \item Even when objectness is explicitly defined, evaluation should employ Average Precision (AP)—including for unknown objects—rather than alternative metrics such as A-OSE or WI.
\end{itemize}

%===========================================================
\section{OSOD-III: An Alternative Formulation}\label{sec:osod-iii}

This section introduces another application formulation of OSOD. Although it has been overlooked in previous studies, we frequently encounter the scenario in practice.
It is free from the fundamental issue of OSOD-II, enabling practical evaluation of methods' performance and probably making the problem easier to solve.

\subsection{OSOD-III: Open at Class Level and  Closed at Super-class Level}\label{sec:def_osodiii}

To motivate the problem we formalize in this section, we begin with two illustrative use cases.

\medskip
\noindent
\textbf{Smartphone insect-detection app}~~
A mobile application that identifies insects from camera images must contend with the vast—and still expanding—number of species. An initial release can cover only a limited subset of known insects. When a user photographs an unfamiliar specimen, the device should flag it as unknown, upload the image to a central server, and allow the model to be updated incrementally. This workflow demands a detector that is closed at the super-class level (``insect'') yet open to previously unseen species within that group.

\medskip
\noindent
\textbf{Traffic-sign detection in advanced driver-assistance systems (ADAS)}~~
Traffic-sign detectors are usually trained on the signs prevalent in the region where a vehicle operates. Signs from other jurisdictions—or newly introduced signs—lie outside this training set, making the detector's response unpredictable. Labeling such signs as unknown is essential both for informing the driver of perceptual uncertainty and for collecting new examples that can be forwarded to a central server for continual learning. Because comprehensive sign inventories are rarely shared with vendors, a detector that is closed at the super-class level (``traffic sign'') but open to unseen sign classes provides a practical solution.

\medskip
These problems are similar to OSOD-II; we want to detect unknown, novel animals. However, unlike OSOD-II, it is unnecessary to consider arbitrary objects as detection targets. In brief, we consider only animal classes; our detector does not need to detect any non-animal object, even if it has been unseen. In other words, we consider the set of object classes closed at the super-class level (i.e., animals) and open at the individual class level under the super-class. 

We call this problem OSOD-III. The differences between OSOD-I, -II, and -III are shown in Fig.~\ref{fig:OSOD_toxm} and Table \ref{tbl:OSOD_toxm}. The problem is formally stated as follows:

\medskip

\noindent
\textbf{OSOD-III}~~ {\em Assume we are given a closed set of object classes belonging to a single super-class. Then, we want to detect and classify objects of these known classes correctly and to detect every unknown class object belonging to the same super-class and classify it as ``unknown.''}

\medskip
\noindent
It is noted that there may be multiple super-classes instead of a single.
In that case, we need only consider the union of the super-classes. For the sake of simplicity, we only consider the case of a single super-class in what follows. 

\subsection{Properties of OSOD-III}\label{sec:prop_osod3}

While the applicability of OSOD-III is narrower than OSOD-II by definition,
OSOD-III has two good properties\footnote{Any OSOD-III problems can be interpreted as OSOD-II. However, it should always be beneficial to formulate it as OSOD-III if possible.}.

One is that OSOD-III is free from the fundamental difficulty of OSOD-II, the dilemma of determining what unknown objects to detect and what to not. Indeed, the judgment is clear with OSOD-III; unknowns belonging to the known super-class should be detected, and all other unknowns should not \footnote{
A key distinction lies in the ``user's intention.''
The fundamental problem with OSOD-II is that there is no way to express what the user wants to detect. In object detection, the term object ''merely signifies a detection target,'' leaving no room for incorporating the user's intent.
In contrast, OSOD-III explicitly provides a means to represent the user's intent. Admittedly, the term `animal'' alone, without any context, may introduce ambiguity in its definition. However, in practice, if necessary, the user can provide a more detailed specification to clarify the meaning. In reality, this ambiguity is unlikely to pose a significant issue.
}.
As a result, OSOD-III no longer suffers from the evaluation difficulty. The clear identification of detection targets enables the computation of AP also for unknown objects.

The other is that detecting unknowns will arguably be easier owing to the similarity between known and unknown classes. In OSOD-II, unknown objects can be arbitrarily dissimilar from known objects. In OSOD-III, known and unknown objects share their super-class, leading to their visual similarity. It should be noted here that what we regard as a super-class is arbitrary; there is no mathematical definition. However, as far as we consider reasonable class hierarchy as in WordNet/ImageNet \citep{WordNet,ImageNet} , we may say that the sub-classes will share visual similarities. 

We emphasize that the formulation of OSOD-III is studied independently of OSOD-II’s limitations; OSOD-III is not intended as a remedy for OSOD-II but instead stands as a self-contained research problem with intrinsic merit.

%===========================================================
\section{Experimental Results}

Based on the above formulation, we evaluate the performance of existing OSOD methods on the proposed OSOD-III scenario.
In the following section, we first introduce our experimental settings to simulate the OSOD-III scenario and then report the evaluation results.

\subsection{Experimental Settings}\label{exp_setting}

\subsubsection{Datasets}\label{sec:dataset}

We use the following three datasets for the experiments: Open Images Dataset v6 \citep{OpenImages}, Caltech-UCSD Birds-200-2011 (CUB200) \citep{CUB200}\footnote{{\color{black}This dataset provides bounding box annotations as well as image labels (\url{https://www.vision.caltech.edu/datasets/cub_200_2011/}).}}, and Mapillary Traffic Sign Dataset (MTSD) \citep{MTSD}. For each, we split classes into known/unknown and images into training/validation/testing subsets as explained below. 
Note that one of the compared methods, ORE \citep{ORE}, needs validation images (i.e., example unknown-class instances), which may be regarded as leakage in OSOD problems. This does not apply to the other methods. {\color{black} Additionally, other datasets can also be used for the OSOD-III formulation, provided that superclass information is defined. Common datasets such as COCO \citep{MSCOCO} and Object365 \citep{Object365} are potential candidates, as they offer some degree of superclass information.}

\medskip \noindent
{\bf Open Images}~~ Open Images \citep{OpenImages} contains 1.9M images of 601 classes of diverse objects with 15.9M bounding box annotations. It also provides the hierarchy of object classes in a tree structure, where each node represents a super-class, and each leaf represents an individual object category. For instance, a leaf {\em Polar Bear} has a parent node {\em Carnivore}. 
We choose two super-classes, {\em Animal} and {\em Vehicle}, in our experiments because of their appropriate numbers of sub-classes, i.e., 96 and 24 in the ``Animal'' and ``Vehicle'' super-class, respectively. We split these sub-classes into known and unknown classes. To mitigate statistical biases, we consider four random splits and select one for a known-class set and the union of the other three for an unknown-class set. 

We construct the training/validation/testing splits of images based on the original splits provided by the dataset. Specifically, we choose the images containing at least one known-class instance from the original training and validation splits. We choose the images containing either at least one known-class instance or at least one unknown-class instance from the original testing split.
For the training images, we keep annotations for the known objects and eliminate all other annotations including unknown objects. It should be noted that there is a risk that those removed objects could be treated as the ``background'' class. For the validation and testing images, we keep the annotations for known and unknown objects and remove all other irrelevant objects.

\medskip \noindent
{\bf CUB200}~~ Caltech-UCSD Birds-200-2011 (CUB200) \citep{CUB200} is a 200 fine-grained bird species dataset. It contains 12K images, for each of which a single box is provided. We split the 200 classes randomly into four splits, each with 50 classes.
We then choose three to form a known-class set and treat the rest as an unknown-class set. 
We construct the training/validation/testing splits similarly to Open Images with two notable exceptions. One is that we create the training/validation/test splits from the dataset's original training/validation splits. This is because the dataset does not provide annotation for the original test split. 
The other is that we remove all the images containing unknown objects from the training splits. This will make the setting more rigorous.

\medskip \noindent
{\bf MTSD}~~ 
Mapillary Traffic Sign Dataset (MTSD) \citep{MTSD} is a dataset of 400 diverse traffic signs from different regions around the world. It contains 52K street-level images with 260K manually annotated traffic sign instances. For the split of known/unknown classes, we consider a practical use case of OSOD-III, where a detector trained using the data from a specific region is used in another region, which might have unknown traffic signs. As the dataset does not provide region information for each image, we divide the 400 traffic sign classes into clusters based on their co-occurrence in the same images. Specifically, we apply normalized graph cut \citep{NormCut1} to obtain three clusters, ensuring any pairs of the clusters share the minimum co-occurrence. We then use the largest cluster as a known-class set (230 classes). Denoting the other two clusters by unknown1 (55) and unknown2 (115), we test three cases, i.e., using either unknown1, unknown2, or their union (unknown1+2) for an unknown-class set. We report the results for the three cases. We create the training/validation/testing splits in the same way as CUB200. 

\medskip

Tables \ref{tbl:oi_data_split}, \ref{tbl:cub200_data_split}, and \ref{tbl:mtsd_data_split} show the resulted splits of each dataset, based on which known/unknown classes are selected, and also those of training/validation/testing images.
See the appendix for more detailed category names contained in each split.

% -------------------------------------------------
% [two column]
\renewcommand{\arraystretch}{1.2}
\begin{table*}[t]
\footnotesize\centering
\caption{Details of the employed class splits for Open Images dataset. We treat one of the four as a known set and the union of the other three as an unknown set. Thus, there are four cases of known/unknown splits, for each of which we report the detection performance in Table \ref{tbl:OpenImages}.}
\label{tbl:oi_data_split}
\scalebox{1.0}{
\begin{tabular}{c||cccc|cccc} \specialrule{0.7pt}{0pt}{0pt}\hline
& \multicolumn{4}{c|}{Animal} & \multicolumn{4}{c}{Vehicle} \\ \cline{2-9}
& Split1 & Split2 & Split3 & Split4 & Split1 & Split2 & Split3 & Split4  \\ \hline
num of known categories & $24$ & $24$ & $24$ & $24$ & $6$ & $6$ & $6$ & $6$ \\
train images & $44,379$ & $38,914$ & $39,039$ & $18,478$ & $43,270$ & $26,860$ & $3,900$ & $6,300$ \\
validation images & $1,104$ & $2,353$ & $1,248$ & $849$ & $1,370$ & $503$ & $178$ & $322$ \\
test images & \multicolumn{4}{c|}{$15,609$} & \multicolumn{4}{c}{$6,991$} \\ \specialrule{0.7pt}{0pt}{0pt}\hline
\end{tabular}
}
\end{table*}
\renewcommand{\arraystretch}{1}
% -------------------------------------------------

% -------------------------------------------------
\renewcommand{\arraystretch}{1.3}
\begin{table}[t]
\footnotesize\centering
\caption{
Details of the employed class splits for Caltech-UCSD Birds-200-2011 (CUB200) dataset. We treat the union of three of the four as known classes and the rest as unknown classes. Each split corresponds to the results shown in Table \ref{tbl:CUB200}.
}
\label{tbl:cub200_data_split}
\scalebox{0.9}{
\begin{tabular}{c||cccc}\specialrule{0.7pt}{0pt}{0pt}\hline
& Split1 & Split2 & Split3 & Split4 \\ \hline
% num of known categories & $150$ & $150$ & $150$ & $150$ \\
num of unknown classes & $50$ & $50$ & $50$ & $50$ \\
train images & $4,109$ & $4,116$ & $4,120$ & $4,120$ \\
validation images & $500$ & $500$ & $500$ & $500$ \\
test images & \multicolumn{4}{c}{$5,794$} \\ \specialrule{0.7pt}{0pt}{0pt}\hline
\end{tabular}
}
\end{table}
\renewcommand{\arraystretch}{1}
% -------------------------------------------------

% -------------------------------------------------
\renewcommand{\arraystretch}{1.3}
\begin{table}[t]
\footnotesize\centering
\caption{Details of the employed class splits for Mapillary Traffic Sign Dataset (MTSD). Each split corresponds to the results shown in Table \ref{tbl:MTSD}. 
}
\label{tbl:mtsd_data_split}
\scalebox{0.9}{
\begin{tabular}{c||ccc}\specialrule{0.7pt}{0pt}{0pt}\hline
& Unknown1 & Unknown2 & Unknown1+2 \\ \hline
num of classes & $55$ & $115$ & $170$ \\
train images & \multicolumn{3}{c}{$13,157$} \\
validation images & \multicolumn{3}{c}{$1,000$} \\
test images & \multicolumn{3}{c}{$3,896$} \\
\specialrule{0.7pt}{0pt}{0pt}\hline
\end{tabular}
}
\end{table}
\renewcommand{\arraystretch}{1}
% -------------------------------------------------

\subsubsection{Evaluation}

As discussed earlier, the primary metric for evaluating object detection performance is average precision (AP) \citep{PBM,VOC}. Although we must use AP for unknown detection, the issue with OSOD-II makes it impractical. OSOD-III is free from the issue, and we can use AP for unknown object detection. Therefore, following the standard evaluation procedure of object detection, we report AP over IoU in the range $[0.50, 0.95]$ for known and unknown object detection. 

\subsection{Compared Methods}\label{sec:compared_methods}
We consider five state-of-the-art OSOD methods \citep{ORE,DropoutSample,VOS,Opendet,OrthogonalDet}. While they are originally developed to deal with OSOD-II, these methods can be applied to OSOD-III without little or no modification. 
{\color{black}
Detection pipelines including methodology design, framework, and dataflow remains unchanged from OSOD-II. 
Accordingly, superclass information is used solely to specify the known/unknown categories that users intend to detect. Neither the superclass name nor its hierarchical information is explicitly incorporated into the training process, whether as supervision or within the loss function.
}

We first summarize five methods and their configurations in our experiments below, followed by the introduction of a simple baseline method that detects unknown objects merely from the outputs of a standard detector.

\medskip \noindent
{\bf ORE (\underline{O}pen Wo\underline{r}ld Object D\underline{e}tector)} \citep{ORE} is initially designed for OWOD; it is capable not only of detecting unknown objects but also of incremental learning. We omit the latter capability and use the former as an open-set object detector. It employs an energy-based method to classify known/unknown; using the validation set, including unknown object annotations, it models the energy distributions for known and unknown objects. To compute AP for unknown objects, we use a detection score that ORE provides. Following the original paper \citep{ORE}, we employ Faster RCNN \citep{FasterRCNN} with a ResNet50 backbone \citep{ResNet} for the base detector.

\medskip \noindent
{\bf DS (\underline{D}ropout \underline{S}ampling)}
\citep{DropoutSample} uses the entropy of class scores to discriminate known and unknown categories. Specifically, during the inference phase, it employs a dropout layer \citep{OrgDropout} right before computing class logits and performs inference $n$ iterations. If the entropy of the average class logits over these iterations exceeds a threshold, the detected instance is assigned to the unknown class. The top-1 class score, calculated from the averaged class logits, is employed as the unknown score for computing unknown AP. The base detector is Faster RCNN with ResNet50-FPN backbone \citep{FPN}. Following the implementation of \citep{Opendet}, we set the number of inference iterations $n$ to 30, the entropy threshold $\gamma_{ds}$ to 0.25, and the dropout layer parameter $p$ to 0.5, respectively.

\medskip \noindent
{\bf VOS (\underline{V}irtual \underline{O}utlier \underline{S}ynthesis)} \citep{VOS} detects unknown objects by treating them as out-of-distribution (OOD) based on an energy-based method. Specifically, it estimates an energy value for each detected instance and judges whether it is known or unknown by comparing the energy with a threshold. We use the energy value to compute unknown AP. We choose Faster RCNN with ResNet50-FPN backbone \citep{FPN}, following the paper.

\medskip \noindent
{\bf OpenDet (\underline{Open}-set \underline{Det}ector)} \citep{Opendet} is the current state-of-the-art on the popular benchmark test designed using PASCAL VOC/COCO (shown in Table \ref{tbl:cat_split_owod}), although the methods' performance is evaluated with inappropriate metrics of A-OSE and WI. OpenDet provides a detection score for unknown objects, which we utilize to compute AP. We use the authors' implementation, which employs Faster RCNN based on ResNet50-FPN for the base detector.

{\color{black}
\medskip \noindent
{\bf OrthogonalDet} \citep{OrthogonalDet} addressed low recall for unknown objects and misclassification into known classes through two key mechanisms: 
1) It promotes decorrelation between objectness prediction and class label prediction, enforcing orthogonality in the feature space.
2) It is based on RandBox \citep{RandBox}, a Fast RCNN-like detector that features the removal of Region Proposal Network (RPN) from Faster RCNN. This modification introduces greater randomness in region proposals, allowing region selection to be independent of the known categories' distribution, thereby achieving high unknown recall. 
We use ResNet50-FPN backbone for this method.
}

\medskip \noindent
{\bf Simple Baselines}~~ 
In addition to these existing methods, we also consider a naive baseline for comparison. It merely uses the class scores that standard detectors predict for each bounding box. It relies on an expectation that unknown-class inputs should result in uncertain class prediction. Thus we look at the prediction uncertainty to judge if the input belongs to known/unknown classes. Specifically, we calculate the ratio of the top-1 and top-2 class scores for each candidate bounding box and compare it with a pre-defined threshold $\gamma$; we regard the input as unknown if it is smaller than $\gamma$ and as known otherwise. We use the sum of the top three class scores for the unknown object detection score. In our experiments, we employ two detectors, FCOS \citep{FCOS} and Faster RCNN \citep{FasterRCNN}. We use ResNet50-FPN as their backbone, following the above methods. For Open Images, we set $\gamma=4.0$ for FCOS and $\gamma=15.0$ for Faster RCNN. For CUB200 and MTSD, we set $\gamma=1.5$ for FCOS and $\gamma=3.0$ for Faster RCNN. We need different thresholds due to the difference in the number of classes and the output layer design, i.e., logistic vs. softmax. Additionally, Faster RCNN could be applied temperature scaling $T$ for the softmax layer. We set $T=1.0$ in our experiments, unless otherwise specified. We report the sensitivity to the choice of $\gamma$ and $T$ in the appendix.

\subsection{Training Details}
We train the models using the SGD optimizer with the batch size of $16$ on 8 A100 GPUs. The number of epochs is 12, 80, and 60 for OpenImages, CUB200, and MTSD, respectively.
We use the initial learning rate of $2.0 \times 10^{-2}$ with momentum $=0.9$ and weight decay $=1.0\times 10^{-4}$. We drop a learning rate by a factor of 10 at $2/3$ and $11/12$ epoch. 
For Open Images and CUB200, we follow a common multi-scale training and resize the input images such that their shorter side is between $480$ and $800$, while the longer side is $1333$ or less. At the inference time, we set the shorter side of input images to $800$ and the longer side to less or equal to $1333$. 
For MTSD, we apply similar scaling strategies to Open Images and CUB200 (i.e., multi-scale training and single-scale testing) but the scaling scheme; namely, the input size is doubled, e.g., the shorter side is between $960$ and $1600$ at training time. This aims to improve detection accuracy for the small-sized objects that frequently appear in MTSD.

We used the publicly available source code for the implementation of ORE\footnote{\url{https://github.com/JosephKJ/OWOD.git}} \citep{ORE}, Dropout Sampling (DS)\footnote{\url{https://github.com/csuhan/opendet2.git}} \citep{DropoutSample}, VOS\footnote{\url{https://github.com/deeplearning-wisc/vos.git}} \citep{VOS}, OpenDet\footnote{\url{https://github.com/csuhan/opendet2.git}} \citep{Opendet}, and {\color{black}OrthogonalDet\footnote{\url{https://github.com/feifeiobama/OrthogonalDet}}}. We used mmdetection\footnote{\url{https://github.com/open-mmlab/mmdetection.git}} \citep{MMDetection} for FCOS \citep{FCOS} and detectron2\footnote{\url{https://github.com/facebookresearch/detectron2}} for Faster RCNN \citep{FasterRCNN} to implement the baseline methods, respectively.

% -------------------------------------------------
\renewcommand{\arraystretch}{1.2}
\begin{table*}[t]
\footnotesize\centering
\caption{
Detection accuracy of known (AP$_{known}$) and unknown objects (AP$_{unk}$) of different methods for Open Images dataset, ``Animal'' and ``Vehicle'' super-classes. ``Split-$n$'' indicates that the classes of Split-$n$ are treated as known classes. ``mean'' is the average of all splits. Orth.Det represents OrthogonalDet.
}
\begin{large}
\label{tbl:OpenImages}
\scalebox{0.63}{
\begin{tabular}{c||cc|cc|cc|cc|cc}\specialrule{0.8pt}{0pt}{0pt}\hline 
& \multicolumn{10}{c}{Animal} \\ \cline{2-11}
& \multicolumn{2}{c|}{Split1} & \multicolumn{2}{c|}{Split2} & \multicolumn{2}{c|}{Split3} & \multicolumn{2}{c|}{Split4} & \multicolumn{2}{c}{mean}\\ \cline{2-11}
& ${\rm AP}_{known}$ & ${\rm AP}_{unk}$ & ${\rm AP}_{known}$ & ${\rm AP}_{unk}$ & ${\rm AP}_{known}$ & ${\rm AP}_{unk}$ & ${\rm AP}_{known}$ & ${\rm AP}_{unk}$ & ${\rm AP}_{known}$ & ${\rm AP}_{unk}$ \\ \hline
ORE \citep{ORE} & $40.4$ & $17.4$ & $34.8$ & $13.0$ & $40.4$ & $19.1$ & $34.8$ & $13.0$ & $37.6\pm 2.8$ & $15.6\pm 2.7$ \\
DS \citep{DropoutSample} & $44.0$ & $19.0$ & $36.8$ & $12.3$ & $43.3$ & $14.0$ & $40.2$ & $14.6$ & $41.1\pm 2.9$ & $15.0\pm 2.5$ \\ 
VOS \citep{VOS} & $39.5$ & $17.5$ & $37.5$ & $13.9$ & $43.1$ & $14.7$ & $37.9$ & $18.1$ & $39.5\pm 2.2$ & $16.0\pm 1.8$ \\ 
OpenDet \citep{Opendet} & $42.4$ & $34.9$ & $23.2$ & $25.8$ & $43.0$ & $37.9$ & $39.0$ & $33.5$ & $36.9\pm 8.1$ & $33.0\pm 4.5$ \\
Orth.Det \citep{OrthogonalDet} & $\textbf{47.0}$ & $11.8$ & $\textbf{40.5}$ & $4.7$ & $\textbf{47.2}$  & $3.1$ & $\textbf{44.5}$ & $15.9$ & $\textbf{44.8} \pm \textbf{3.1}$ & $8.9 \pm 6.0$\\
FCOS \citep{FCOS} & $35.0$ & $\textbf{44.4}$ & $30.8$ & $\textbf{35.6}$ & $32.6$ & $\textbf{43.7}$ & $22.6$ & $\textbf{43.6}$ & $30.3\pm 4.7$ & $\textbf{41.8}\pm \textbf{3.6}$ \\
Faster RCNN \citep{FasterRCNN} & $41.8$ & $36.9$ & $34.0$ & $29.5$ & $39.7$ & $37.7$ & $35.5$ & $37.0$ & $37.8\pm 3.1$ & $35.3\pm 3.9$ \\ 
\hline

& \multicolumn{10}{c}{Vehicle} \\ \cline{2-11}
& \multicolumn{2}{c|}{Split1} & \multicolumn{2}{c|}{Split2} & \multicolumn{2}{c|}{Split3} & \multicolumn{2}{c|}{Split4} & \multicolumn{2}{c}{mean}\\ \cline{2-11}
& ${\rm AP}_{known}$ & ${\rm AP}_{unk}$ & ${\rm AP}_{known}$ & ${\rm AP}_{unk}$ & ${\rm AP}_{known}$ & ${\rm AP}_{unk}$ & ${\rm AP}_{known}$ & ${\rm AP}_{unk}$ & ${\rm AP}_{known}$ & ${\rm AP}_{unk}$ \\ \hline
ORE \citep{ORE} & $46.9$ & $0.5$ & $35.0$ & $0.1$ & $25.0$ & $0.2$ & $27.7$ & $0.3$ & $33.7\pm 8.5$ & $0.3\pm 0.1$ \\
DS \citep{DropoutSample} & $52.6$ & $0.5$ & $40.7$ & $2.3$ & $31.9$ & $6.5$ & $35.1$ & $1.4$ & $40.1\pm 7.9$ & $2.7\pm 2.3$ \\ 
VOS \citep{VOS} & $53.2$ & $7.4$ & $41.9$ & $7.1$ & $\textbf{32.8}$ & $9.4$ & $35.7$ & $12.6$ & $40.9\pm 7.8$ & $9.1\pm 2.2$ \\ % without validation
OpenDet \citep{Opendet} & $50.6$ & $10.2$ & $40.4$ & $12.5$ & $30.2$ & $15.9$ & $33.6$ & $19.0$ & $38.7\pm 7.8$ & $14.4\pm 3.3$ \\ 
Orth.Det \citep{OrthogonalDet} & $\textbf{56.3}$ & $2.5$ & $41.3$ & $1.2$ & $29.7$ & $16.1$ & $34.6$ & $2.5$ & $\textbf{40.5}\pm \textbf{11.6}$ & $5.6\pm 7.0$\\
FCOS \citep{FCOS} & $49.6$ & $\textbf{14.2}$ & $32.7$ & $14.6$ & $19.2$ & $\textbf{24.7}$ & $21.4$ & $\textbf{21.3}$ & $30.7\pm 12.0$ & $\textbf{18.7}\pm \textbf{4.5}$ \\ 
Faster RCNN \citep{FasterRCNN} & $51.0$ & $10.5$ & $\textbf{42.0}$ & $\textbf{15.2}$ & $31.0$ & $22.1$ & $\textbf{35.7}$ & $20.2$ & $39.9\pm 8.7$ & $17.0\pm 5.2$ \\ 
\specialrule{0.8pt}{0pt}{0pt}\hline
\end{tabular}
}
\end{large}
\end{table*}
\renewcommand{\arraystretch}{1}
% -------------------------------------------------
% -------------------------------------------------
\renewcommand{\arraystretch}{1.2}
\begin{table*}[t]
\footnotesize\centering
\caption{Detection accuracy for CUB200 \citep{CUB200}. See Table \ref{tbl:OpenImages} for notations.}
\begin{large}
\label{tbl:CUB200}
\scalebox{0.64}{
\begin{tabular}{c||cc|cc|cc|cc|cc}\specialrule{0.8pt}{0pt}{0pt}\hline 
& \multicolumn{2}{c|}{Split1} & \multicolumn{2}{c|}{Split2} & \multicolumn{2}{c|}{Split3} & \multicolumn{2}{c|}{Split4} & \multicolumn{2}{c}{mean}\\ \cline{2-11}
& ${\rm AP}_{known}$ & ${\rm AP}_{unk}$ & ${\rm AP}_{known}$ & ${\rm AP}_{unk}$ & ${\rm AP}_{known}$ & ${\rm AP}_{unk}$ & ${\rm AP}_{known}$ & ${\rm AP}_{unk}$ & ${\rm AP}_{known}$ & ${\rm AP}_{unk}$ \\ \hline
ORE \citep{ORE} & $51.3$ & $18.1$ & $53.6$ & $21.8$ & $54.4$ & $17.7$ & $53.6$ & $21.6$ & $53.2\pm 1.3$ & $19.8\pm 2.2$ \\
DS \citep{DropoutSample} & $61.7$ & $19.6$ & $61.2$ & $22.2$ & $62.8$ & $22.2$ & $60.4$ & $21.8$ & $61.5\pm 0.9$ & $21.5\pm 1.1$ \\
VOS \citep{VOS} & $59.7$ & $8.1$ & $59.5$ & $9.1$ & $60.5$ & $8.1$ & $57.7$ & $9.5$ & $59.4\pm 1.0$ & $8.7\pm 0.6$ \\ % follow paper
OpenDet \citep{Opendet} & $63.9$ & $23.1$ & $63.6$ & $\textbf{30.0}$ & $63.9$ & $\textbf{26.3}$ & $61.6$ & $\textbf{28.6}$ & $63.3\pm 1.1$ & $\textbf{27.0}\pm \textbf{3.0}$ \\
Orth.Det \citep{OrthogonalDet} & $\textbf{64.9}$ & $\textbf{27.2}$ & $\textbf{67.2}$ & $26.9$ & $\textbf{67.7}$ & $20.9$ & $\textbf{66.0}$ & $23.8$ & $\textbf{66.5} \pm \textbf{1.3}$ & $24.7 \pm 3.0$ \\
FCOS \citep{FCOS} & $55.0$ & $23.0$ & $55.2$ & $26.1$ & $50.6$ & $25.0$ & $53.0$ & $24.6$ & $53.5\pm 2.1$ & $24.7\pm 1.3$ \\
Faster RCNN \citep{FasterRCNN} & $62.0$ & $21.6$ & $62.7$ & $26.2$ & $63.2$ & $24.0$ & $60.8$ & $24.8$ & $62.2\pm 1.0$ & $24.2\pm 1.9$ \\
\specialrule{0.8pt}{0pt}{0pt}\hline
\end{tabular}
}
\end{large}
\end{table*}
% -------------------------------------------------
% -------------------------------------------------
\renewcommand{\arraystretch}{1.3}
\begin{table*}[t]
\footnotesize\centering
\caption{Detection accuracy for MTSD \citep{MTSD}. K, U1, and U2 stand for the splits of Known, Unknown1, and Unknown2, respectively.
}
\label{tbl:MTSD}
\scalebox{1.0}{
\begin{tabular}{c||c|c|c|c|c}\specialrule{0.8pt}{0pt}{0pt}\hline  
& \multicolumn{1}{c|}{K} & \multicolumn{1}{c|}{U1} & \multicolumn{1}{c|}{U2} & \multicolumn{1}{c|}{U1+2} & \multicolumn{1}{c}{mean} \\ \cline{2-6}
& \multicolumn{1}{c|}{${\rm AP}_{known}$} & \multicolumn{4}{c}{${\rm AP}_{unk}$} \\ \hline
ORE \citep{ORE} & $41.2$ & $0.4$ & $0.2$ & $0.7$ & $0.4\pm 0.3$ \\
DS \citep{FasterRCNN} & $50.4$ & $4.5$ & $3.4$ & $7.5$ & $5.1\pm 1.7$ \\ 
VOS \citep{VOS} & $49.1$ & $4.6$ & $2.9$ & $6.5$ & $4.7\pm 1.5$ \\ % follow paper
OpenDet \citep{Opendet}        & $\textbf{51.8}$ & $\textbf{8.7}$ & $\textbf{6.7}$ & $\textbf{14.2}$ & $\textbf{9.9}\pm \textbf{3.9}$ \\
Orth.Det \citep{OrthogonalDet} & $49.3$ & $3.9$ & $1.5$ & $4.4$ & $3.3\pm 1.6$ \\
FCOS \citep{FCOS}              & $41.7$ & $3.8$ & $3.3$ & $6.2$ & $4.4\pm 1.6$ \\
Faster RCNN \citep{FasterRCNN} & $50.0$ & $2.5$ & $2.3$ & $4.4$ & $3.1\pm 1.2$ \\ 
\specialrule{0.8pt}{0pt}{0pt}\hline
\end{tabular}
}
\end{table*}
\renewcommand{\arraystretch}{1}
% -------------------------------------------------

\subsection{Results}\label{sec:results}

\subsubsection{Main Results}\label{sec:main_results}

Tables \ref{tbl:OpenImages}, \ref{tbl:CUB200}, and \ref{tbl:MTSD}
present the results for Open Images, CUB200, and MTSD. They all show mAP for known-class objects (${\rm AP}_{known}$) and AP for unknown objects (${\rm AP}_{unk}$).

We can see a consistent trend irrespective of datasets. For all methods, while mAP for known classes is high, AP for the unknown class is low. 
{\color{black}
Especially for the insufficient ${\rm AP}_{unk}$ scores on ``Vehicle'' dataset in Table~\ref{tbl:OpenImages}, we hypothesize two factors. First, DS and ORE, being among the earliest OSOD models, struggle with distinguishing unknown objects and result in low unknown recall.
Second, with only six known categories, the ``Vehicle'' dataset provides limited supervision, causing DS and ORE to overfit rather than generalize the superclass concept.
}
Overall, a comparison of techniques shows that OpenDet exhibits superior performance in detecting the unknown class compared to other existing methods. 
The performance of OrthogonalDet in unknown object detection is comparable to or worse than that of other methods across datasets. 
However, it is noteworthy that our naïve baseline demonstrates comparable or even superior performance. This is surprising, considering that they do not require additional training or mechanism dedicated to unknown detection. 

Nonetheless, even the best-performing methods achieve only modest accuracy in unknown detection, rendering them unsuitable for practical application. While reasonable accuracy is achieved in the ``Animal'' superclasses of Open Images, the evaluation on the more realistic MTSD results in significantly lower accuracy. This suggests that there is substantial room for further research in OSOD(-III). This is especially true considering that our baseline has achieved the highest level of accuracy, indicating significant potential for improvement.

\subsubsection{Superclass Specification at Different Levels}\label{sec:diff_level_superclass}
To explore the impact of superclass specification at different hierarchical levels, we conducted additional experiments using superclasses from the multi-level hierarchy of the Open Images dataset.
We selected ``Carnivore'' and ``Land Vehicle'' from the Open Images dataset as superclasses at different hierarchical levels, being subcategories of ``Animal'' and ``Vehicle,'' respectively. 
We followed the same process used when creating the ``Animal'' and ``Vehicle'' benchmarks of the Open Images dataset. Specifically, we split the target categories into two splits (i.e., eight known and unknown categories) and created train/validation/test datasets. Table~\ref{tbl:carnivore_landvehicle_data_split} presents the details of resulting dataset.
We then evaluated the detection performance on each split.

The evaluation results are shown in Table.~\ref{tbl:subsuperclass}. We use OpenDet and Faster RCNN as representative methods, as they demonstrated strong performance in our main experiments on Open Images dataset presented in the main paper.
We observe that our naive baseline, Faster RCNN, achieves comparable performance on both known and unknown classes in the ``Carnivore'' dataset, exhibiting a similar trend to that of the ``Animal'' superclass. 
In the ``Land Vehicle'' dataset, Faster RCNN outperforms OpenDet in detecting known classes but underperforms in detecting unknown classes, highlighting a trade-off between the two.
We speculate that as the specified superclass becomes a lower-level concept (i.e., closer to a fine-grained category), the limited domains of known and unknown objects may provoke detectors to focus more on closed-set learning of known categories rather than generalizing to the superclass. Consequently, this could lead to overfitting on the known classes.

% -------------------------------------------------
\renewcommand{\arraystretch}{1.3}
\begin{table}[t]
\footnotesize\centering
\caption{
Details of the employed class splits for ``Carnivore'' and ``Land Vehicle'' datasets. 
}
\label{tbl:carnivore_landvehicle_data_split}
\scalebox{0.88}{
\begin{tabular}{c||cc|cc}\specialrule{0.7pt}{0pt}{0pt}\hline
& \multicolumn{2}{c|}{Carnivore} & \multicolumn{2}{c}{Land Vehicle} \\ \hline
& Split1 & Split2 & Split1 & Split2 \\ \hline
num of unknown classes & $8$ & $8$ & $8$ & $8$ \\
train images & $16,097$ & $24,850$ & $34,860$ & $26,699$ \\
validation images & $2,218$ & $2,218$ & $1,268$ & $1,268$ \\
test images & \multicolumn{2}{c|}{$6,753$} & \multicolumn{2}{c}{$3,819$} \\ 
\specialrule{0.7pt}{0pt}{0pt}\hline
\end{tabular}
}
\end{table}
\renewcommand{\arraystretch}{1}
% -------------------------------------------------

% -------------------------------------------------
% [Details of OpenImages results]
\renewcommand{\arraystretch}{1.2}
\begin{table*}[ht]
\footnotesize\centering
\caption{
Detection accuracy of known (AP$_{known}$) and unknown objects (AP$_{unk}$) of ``Carnivore'' and ``Land Vehicle''. 
See Table \ref{tbl:OpenImages} for notations.
}
\begin{large}
\label{tbl:subsuperclass}
\scalebox{0.7}{
\begin{tabular}{c||cc|cc|cc}\specialrule{0.8pt}{0pt}{0pt}\hline 
& \multicolumn{6}{c}{Carnivore} \\ \cline{2-7}
& \multicolumn{2}{c|}{Split1} & \multicolumn{2}{c|}{Split2} & \multicolumn{2}{c}{mean}\\ \cline{2-7}
& ${\rm AP}_{known}$ & ${\rm AP}_{unk}$ & ${\rm AP}_{known}$ & ${\rm AP}_{unk}$ & ${\rm AP}_{known}$ & ${\rm AP}_{unk}$ \\ \hline
OpenDet \citep{Opendet} & $30.9$ & $\textbf{40.8}$ & $47.5$ & $16.2$ & $39.2\pm 8.3$ & $28.5\pm 12.3$\\
Faster RCNN \citep{FasterRCNN} & $\textbf{31.6}$ & $39.3$ & $\textbf{54.4}$ & $\textbf{21.5}$ & $\textbf{43.0}\pm \textbf{11.4}$ & $\textbf{30.4}\pm \textbf{8.9}$\\
\hline\hline

& \multicolumn{6}{c}{Land Vehicle} \\ \cline{2-7}
& \multicolumn{2}{c|}{Split1} & \multicolumn{2}{c|}{Split2} & \multicolumn{2}{c}{mean}\\ \cline{2-7}
& ${\rm AP}_{known}$ & ${\rm AP}_{unk}$ & ${\rm AP}_{known}$ & ${\rm AP}_{unk}$ & ${\rm AP}_{known}$ & ${\rm AP}_{unk}$ \\ \hline
OpenDet \citep{Opendet} & $43.6$ & $\textbf{11.0}$ & $42.6$ & $\textbf{17.6}$ & $43.1 \pm 0.5$ & $\textbf{14.3} \pm \textbf{3.3}$ \\
Faster RCNN \citep{FasterRCNN} & $\textbf{48.4}$ & $2.3$ & $\textbf{45.3}$ & $8.9$ & $\textbf{46.8} \pm \textbf{1.6}$ & $5.6 \pm 3.3$ \\
\hline
\specialrule{0.8pt}{0pt}{0pt}\hline
\end{tabular}
}
\end{large}
\end{table*}
\renewcommand{\arraystretch}{1}

\subsubsection{Results of A-OSE and WI}
In addition to ${\rm AP}_{known}$ and ${\rm AP}_{unk}$, we report absolute open-set error (A-OSE) and wilderness impact (WI), the metrics widely used in previous studies. 
Table~\ref{tbl:wi_aose_mean} shows those for the compared methods on the same test data.
Recall that i) A-OSE and WI measure only detectors' performance of known object detection; and ii) they evaluate detectors' performance at a single operating point. 
Table~\ref{tbl:wi_aose_mean} shows the results at the operating points chosen in the previous studies, i.e., confidence score $>0.05$ for A-OSE and the recall (of known object detection) $=0.8$ for WI, respectively. 
The results show that OpenDet and Faster RCNN achieve comparable performance on both metrics. 
FCOS performs worse, but this is not necessarily true at different operating points, as shown in Fig.~\ref{fig:aose_wi_plot} of the main paper.
We can also see from the results a clear inconsistency between the A-OSE/WI and ${\rm AP}$s. 
For instance, Faster RCNN is inferior to ORE in both the A-OSE and WI metrics (i.e., $6,382\pm 206$ vs. $4,849\pm 206$ on A-OSE), whereas it achieves much better ${\rm AP}_{known}$ and ${\rm AP}_{unk}$ than ORE, as shown in Table~\ref{tbl:CUB200}. Such inconsistency demonstrates that A-OSE and WI are unsuitable performance measures for OSOD-II/III, as discussed in Sec~\ref{sec:wi_aose_defacts}.

% -------------------------------------------------
\renewcommand{\arraystretch}{1.5}
\begin{table*}[t]
\footnotesize\centering
\caption{
A-OSE and WI of the compared methods in the experiment of each dataset. The same experimental settings as Table~\ref{tbl:OpenImages}–\ref{tbl:MTSD} are used. The reported values are the averages across all splits. Results for individual splits can be found in the appendix. }
\label{tbl:wi_aose_mean}
\scalebox{0.75}{
\begin{tabular}{c||cc|cc|cc|cc}
\specialrule{0.8pt}{0pt}{0pt}\hline
& \multicolumn{2}{c|}{Animal} & \multicolumn{2}{c|}{Vehicle} & \multicolumn{2}{c|}{CUB200} & \multicolumn{2}{c}{MTSD} \\ \cline{2-9}
Method & A-OSE & WI & A-OSE & WI & A-OSE & WI & A-OSE & WI \\ \hline
ORE\citep{ORE}         & $22,268 \pm 2,848$ & $39.7 \pm 6.7$   & $4,514 \pm 1,323$ & $24.9 \pm 6.0$   & $4,849 \pm 206$   & $22.1 \pm 2.0$   & $2,348 \pm 827$   & $8.0 \pm 3.3$ \\
DS\citep{DropoutSample} & $38,776 \pm 6,185$ & $50.1 \pm 9.4$   & $11,025 \pm 4,204$ & $37.1 \pm 11.0$  & $\textbf{3,363} \pm \textbf{125}$   & $\textbf{19.1} \pm \textbf{2.1}$   & $2,495 \pm 899$   & $10.2 \pm 3.7$ \\
VOS\citep{VOS}         & $22,024 \pm 6,714$ & $46.1 \pm 11.2$  & $\textbf{2,083} \pm \textbf{611}$   & $\textbf{23.8} \pm \textbf{9.5}$   & $4,415 \pm 498$   & $20.6 \pm 1.6$   & $2,175 \pm 1,013$ & $9.5 \pm 4.5$ \\
OpenDet\citep{Opendet} & $25,252 \pm 1,585$ & $41.1 \pm 10.7$  & $7,130 \pm 2,502$ & $31.9 \pm 12.2$  & $4,539 \pm 167$   & $20.1 \pm 2.2$   & $\textbf{1,245} \pm \textbf{579}$   & $7.8 \pm 3.9$ \\
Orth.Det\citep{OrthogonalDet} & $310,335 \pm 49,663$ & $\textbf{38.4} \pm \textbf{9.9}$   & $74,416 \pm 26,789$ & $33.4 \pm 8.4$   & $20,134 \pm 2,832$ & $21.8 \pm 1.24$  & $55,063 \pm 24,374$ & $\textbf{3.9} \pm \textbf{1.8}$ \\
FCOS\citep{FCOS}       & $39,166 \pm 8,053$ & $45.5 \pm 9.6$   & $14,121 \pm 5,558$ & $37.6 \pm 10.9$  & $17,988 \pm 2,553$ & $25.6 \pm 1.6$   & $7,989 \pm 3,628$  & $8.5 \pm 3.8$ \\
Faster RCNN\citep{FasterRCNN} & $\textbf{14,438} \pm \textbf{2,314}$ & $40.4 \pm 13.8$ & $5,444 \pm 1,956$  & $33.7 \pm 16.2$ & $6,382 \pm 206$   & $22.7 \pm 3.7$   & $1,897 \pm 868$   & $8.8 \pm 4.0$ \\
\specialrule{0.8pt}{0pt}{0pt}\hline
\end{tabular}
}
\end{table*}
\renewcommand{\arraystretch}{1}
% -------------------------------------------------

\subsection{Analysis on Failure Cases}\label{sec:analysis}

% -------------------------------------------------
\begin{figure*}[t]
\centering
\includegraphics[width=1.0\linewidth]{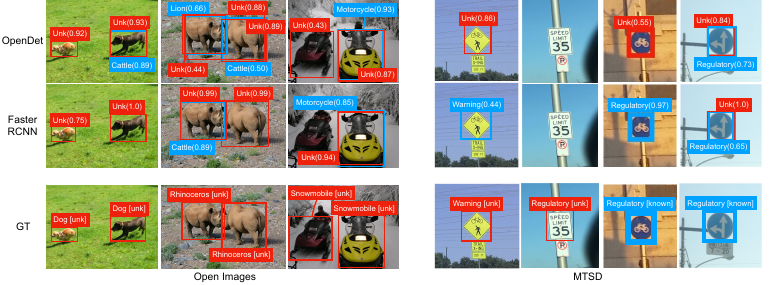}
\caption{Example outputs of OpenDet \citep{Opendet} and our baseline method with Faster RCNN \citep{FasterRCNN} for Open Images with Animal and Vehicle super-classes and MTSD, respectively. Red and blue boxes indicate detected unknown-class and known-class objects, respectively; ``Unk'' means ``unknown''.
}
\label{fig:vis_oi_mtsd}
\end{figure*}
% -------------------------------------------------

\subsubsection{Overlapped Detections between Known and Unknown}\label{sec:analysis_overlap}
To understand the reasons behind the low accuracy in unknown detection, we analyzed failure cases. This analysis revealed that, in addition to a certain degree of false negatives in unknown detection, there is a significant confusion between unknown and known classes. Some examples are shown in Fig.~\ref{fig:vis_oi_mtsd}, with more detailed data deferred to the appendix. As shown in Fig.~\ref{fig:vis_oi_mtsd}, it was found that bounding boxes (BB) for unknown and known classes often overlap, contributing to this confusion.

The methods we evaluated in our experiments all employ non-maximum suppression (NMS). If NMS were applied across two BBs of unknown and a known class (i.e., eliminating the lower-scoring one), such issues might be mitigated. In our experiments, NMS treated the unknown class as equivalent to one of the known classes. In standard object detection procedures, NMS is applied separately for each class, meaning that known and unknown were subjected to NMS independently. It may be beneficial to apply NMS across unknown and known classes. Therefore, we conducted experiments to explore this approach.

\begin{figure*}[t]
\centering
\includegraphics[width=0.8\linewidth]{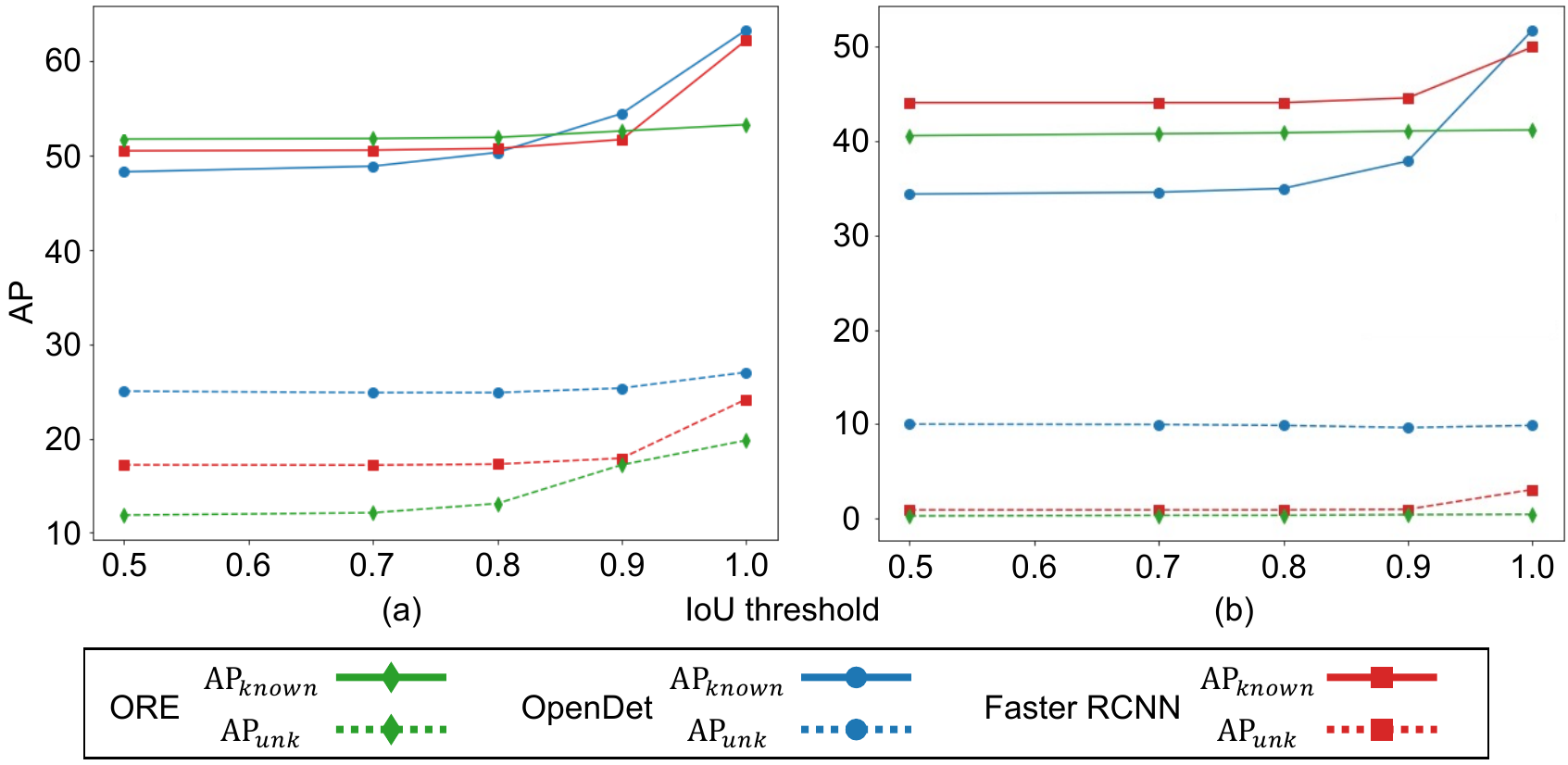}
\caption{
Detection accuracy at various IoU thresholds for NMS between known and unknown predictions: mAP for known classes and AP for unknown. The results for (a) CUB200 and (b) MTSD.
}
\label{fig:kuNMS}
\end{figure*}

Figure~\mbox{\ref{fig:kuNMS}} shows the mAP for the known category predictions and AP for the unknown, evaluated at varying IoU thresholds for NMS. In Fig~\mbox{\ref{fig:kuNMS}}, an IoU threshold of 1.0 represents results obtained without NMS between known and unknown predictions. Results for other values demonstrate the effect of NMS on these predictions. It is clear that aggressive NMS reduces APs for both categories. 

This observation yields two insights: i) predicted known and unknown BBs do frequently overlap, and ii) the scores of these BBs do not consistently reflect prediction accuracy. Ideally, when BBs overlap, the one with the highest score should correspond to the correct prediction. However, in our results, BBs, which incorrectly classify instances as either known or unknown, often receive higher scores than those making accurate classifications. In summary, while the detectors show competence in identifying unknown instances, they regularly misidentify between known and unknown instances. These findings will offer useful insights for the advancement of more effective OSOD methods.

% -------------------------------------------------
% -------------------------------------------------

\subsubsection{Feature Space Visualization}\label{sec:analysis_featurespace}
For more analysis on the confusion between known and unknown, we performed t-SNE visualization of latent features for both detectors trained under OSOD-II and OSOD-III scenarios. 
For the OSOD-II setting, we utilized the VOC-COCO dataset \citep{ORE,Opendet}, which defines 20 PASCAL-VOC classes as known categories and 60 non-PASCAL-VOC classes as the unknown category. For the OSOD-III setting, we employed four of our proposed benchmarks.
Specifically, we first sampled RoI features from the penultimate layer before the final classification head. Only RoI features with an IoU of 0.8 or higher with any ground truth (GT) instance were selected. From these, 30 samples were randomly chosen for t-SNE analysis. For visualization, 10 known classes and one unknown class were randomly sampled.

\begin{figure*}[h]
\centering
\includegraphics[width=1.0\linewidth]{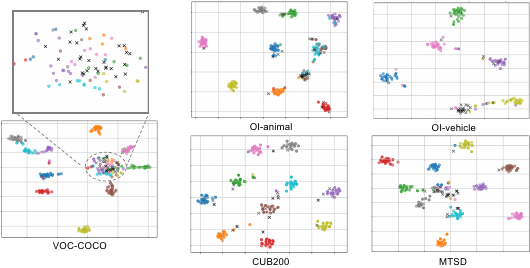}
\caption{t-SNE visualization of the feature space of OpenDet trained under OSOD-II (VOC-COCO) and OSOD-III (four proposed benchmarks) scenarios. Latent features from known classes (colored circles) and unknown class (black crosses) are randomly sampled.}
\label{fig:feature_space}
\end{figure*}

Figure~\ref{fig:feature_space} presents the results. 
We can observe that known features are reasonably well-separated in the feature space, highlighting the strong performance on $\rm{AP}_{known}$ in the main experiments. In the VOC-COCO dataset (i.e., OSOD-II formulation), the unknown class tends to map to regions of the feature space where various known categories intermingle, indicating relatively low similarity to known clusters.
For the four proposed datasets (i.e., OSOD-III formulation), unknown features are more likely to be mapped near known clusters rather than being concentrated in a distinct region as a separate unknown class. This observation is reasonable, as the known and unknown classes share the same superclass in OSOD-III.
The overlap between these two classes in the feature space supports the credibility of our analysis of failure cases, that is, OSOD-III is more prone to misidentification between known and unknown classes rather than discovery of unknown objects.

% -------------------------------------------------
% -------------------------------------------------

%===========================================================
\section{Related Work}\label{sec:relatedwork}

\subsection{Open-set Recognition}
For the safe deployment of neural networks, open-set recognition (OSR) has attracted considerable attention.
The task of OSR is to accurately classify known objects and simultaneously detect unseen objects as unknown. Scheirer \etal \citep{org_OSR} first formulated the problem of OSR, and many following studies have been conducted so far \citep{OpenMax,GAN_OpenMax,Counterfactual_OSR,CGD_OSR,C2AE,DOC,ClosedSet_ICLR,PROSER}.

The work of Bendale and Boult \citep{OpenMax} is the first to apply deep neural networks to OSR. They use outputs from the penultimate layer of a network to calibrate its prediction scores.
Several studies \citep{GAN_OpenMax,Counterfactual_OSR,OpenGAN} found generative models are effective for OSR, where unseen-class images are synthesized and used for training.
Another line of OSR studies focuses on a reconstruction-based method using latent features \citep{SparceOSR,CROSR}, class conditional auto-encoder \citep{C2AE}, and conditional gaussian distributions \citep{CGD_OSR}.

\subsection{Open-set Object Detection}
We can categorize existing open-set object detection (OSOD) problems into two scenarios, OSOD-I and -II, according to their different interest in unknown objects, as we have discussed in this paper.

\medskip \noindent
{\bf OSOD-I}~ Early studies treat OSOD as an extension of OSR problem \citep{DropoutSample,Uncertain_ObjDet,Overlooked}. They aim to correctly detect every known object instance and avoid misclassifying any unseen object instance into known classes. Miller \etal \citep{DropoutSample} first utilize multiple inference results through dropout layers \citep{OrgDropout} to estimate the uncertainty of the detector's prediction and use it to avoid erroneous detections under open-set conditions. Dhamija \etal \citep{Overlooked} investigate how modern CNN detectors behave in an open-set environment and reveal that the detectors detect unseen objects as known objects with a high confidence score. For the evaluation, researchers have employed A-OSE \citep{DropoutSample} and WI \citep{Overlooked} as the primary metrics to measure the accuracy of detecting known objects. They are designed to measure how frequently a detector wrongly detects and classifies unknown objects as known objects.

\medskip \noindent
{\bf OSOD-II}~ More recent studies have moved in a more in-depth direction, where they aim to correctly detect/classify every object instance not only with the known class but also with the unknown class. This scenario is often considered a part of open-world object detection (OWOD) \citep{ORE,OW-DETR,RevisitOWOD,ORDER,UC-OWOD}. In this case, the detection of unknown objects matters since it considers updating the detectors by collecting unknown classes and using them for retraining. Joseph \etal \citep{ORE} first introduces the concept of OWOD and establishes the benchmark test. Many subsequent works have strictly followed this benchmark and proposed methods for OSOD. OW-DETR \citep{OW-DETR} introduces a transformer-based detector (i.e., DETR \citep{DETR,DDETR}) for OWOD and improves the performance. Han \etal \citep{Opendet} propose OpenDet and pay attention to the fact that unknown classes are distributed in low-density regions in the latent space. They then perform contrastive learning to encourage intra-class compactness and inter-class separation of known classes, leading to performance gain. Similarly, Du \etal \citep{VOS} synthesize virtual unseen samples from the decision boundaries of gaussian distributions for each known class. 
Wu \etal \citep{UC-OWOD} propose a further challenging task to distinguish unknown instances as multiple unknown classes.

\subsection{Hierarchical Novelty Detection}
Recent studies have explored hierarchical novelty detection (HND), which conceptually resembles OSOD-III.

Lee et al. \citep{HND} introduced a hierarchical classification framework for novelty detection, leveraging a taxonomy of known classes to identify the most relevant superclass for novel objects. 
More recently, Pyakurel and Yu \citep{HND_ICML} proposed fine-grained evidence allocation for HND, improving detection precision by introducing virtual novel classes at each non-leaf level and enabling a structured, evidence-based multi-class classification approach. 

Although HND and OSOD-III share conceptual similarities, they differ in three key aspects. 1) HND focuses on image-level classification, whereas OSOD-III operates at the instance-level for object detection. 2) HND aims to identify the closest superclass for detected unknown objects, within the hierarchical taxonomy. In contrast, OSOD-III does not require superclass inference, as the superclass is predefined by the user. This reduces task complexity, making OSOD-III feasible even for instance-level tasks. 3) To facilitate superclass inference, HND explicitly constructs a category hierarchy (e.g., WordNet \citep{WordNet}) as prior knowledge, providing models with a structured taxonomy for classification. OSOD-III does not necessitate such a hierarchy; instead, detectors are supplied with a list of known categories from which they learn the corresponding superclass.

%===========================================================
\section{Conclusion}

In this paper, we have studied the problem of open-set object detection (OSOD). Initially, we categorized existing problem formulations in the literature into two types: OSOD-I and OSOD-II. We then highlighted the inherent difficulties in OSOD-II, the most widely studied formulation, where the primary issue is identifying which unknown objects to detect. This ambiguity renders practical evaluation of OSOD-II problematic.

Subsequently, we introduced a novel OSOD formulation, OSOD-III, which focuses on detecting unknown objects that belong to the same super-class as known objects. This perspective, previously neglected in the field, is of significant practical relevance. We demonstrated that OSOD-III is not subject to the issues plaguing OSOD-II, enabling the effective evaluation of methodologies using the standard AP metric. We also established benchmark tests for OSOD-III and evaluated various methods, including the current state-of-the-art. Our primary finding is that existing methods achieve only modest performance, falling short of practical application in real-world scenarios. Our analyses revealed that the primary challenge lies not in detecting unknown instances, but in differentiating known from unknown instances. Anticipating further advancements, we hope our insights will contribute to future developments in this area.

\medskip \noindent{\bf Acknowledgments}

This work was supported by JST SPRING, Grant Number JPMJSP2114.

 %===========================================================
\begin{appendices}

\section{Access to Our Dataset}

Our datasets can be accessed at \url{https://github.com/rsCPSyEu/OSOD-III.git}.

%===========================================================
\section{Details of the Datasets}

We used three datasets in our experiments, i.e., Open Images dataset \citep{OpenImages}, Caltech-UCSD Birds-200-2011 (CUB200) \citep{CUB200}, and Mapillary Traffic Sign Dataset (MTSD) \citep{MTSD}. 
Tables~\ref{tbl:category_details_oi}, \ref{tbl:category_details_cub200}, and \ref{tbl:category_details_mtsd} provide lists of the classes for each split. Please check Sec~\ref{sec:dataset} in the main paper as well.

%===========================================================
% -------------------------------------------------
\renewcommand{\arraystretch}{1.3}
\begin{table*}[h]
\centering
\caption{
Classes contained in the employed splits for Open Images \citep{OpenImages} with the super-classes ``Animal'' (first column) and ``Vehicle'' (second column), respectively.
}
\label{tbl:category_details_oi}
\scalebox{0.8}{
\begin{tabular}{c||c|c}\specialrule{0.7pt}{0pt}{0pt}\hline
& Animal & Vehicle \\ \hline
\begin{tabular}{c} 
Split1\\
(24/6)
\end{tabular} &
\begin{tabular}{c} 
Starfish / Deer / Tick / Lynx / \\
Monkey / Squirrel / Koala / Fox / \\
Spider / Scorpion / Rabbit / Hamster / \\ 
Woodpecker / Snail / Brown bear / Polar bear /\\
Lion / Bull / Shrimp / Panda / \\
Chicken / Sparrow / Cattle / Lobster
\end{tabular} & 
\begin{tabular}{c} 
Bicycle / Golf cart / \\
Van / Taxi / \\
Airplane / Motorcycle
\end{tabular} \\ \hline

\begin{tabular}{c} 
Split2\\
(24/6)
\end{tabular} &
\begin{tabular}{c} 
Sea lion / Mule / Lizard / Raccoon / \\
Butterfly / Hippopotamus / Kangaroo / Frog / \\
Harbor seal / Red panda / Antelope / Ant / \\
Sheep / Dog / Magpie / Teddy bear / \\
Oyster / Otter / Seahorse / Caterpillar / \\
Worm / Zebra / Jaguar (Animal) / Rays and skates 
\end{tabular} & 
\begin{tabular}{c} 
Train / Truck / \\
Barge / Gondola / \\
Rocket / Bus
\end{tabular} \\ \hline

\begin{tabular}{c} 
Split3\\
(24/6)
\end{tabular} &
\begin{tabular}{c} 
Tortoise / Skunk / Blue jay / Rhinoceros / \\
Turkey / Falcon / Dinosaur / Bat (Animal) / \\
Squid / Giraffe / Owl / Armadillo / \\
Swan / Duck / Goose / Camel / \\
Horse / Tiger / Goldfish / Cat / \\
Shark / Parrot / Leopard / Goat
\end{tabular} & 
\begin{tabular}{c} 
Submarine / Jet ski / \\
Unicycle / Snowmobile / \\
Cart / Tank
\end{tabular} \\ \hline

\begin{tabular}{c} 
Split4\\
(24/6)
\end{tabular} &
\begin{tabular}{c} 
Dragonfly / Ladybug / Raven / Penguin / \\
Hedgehog / Mouse / Snake / Jellyfish / \\
Porcupine / Ostrich / Elephant / Dolphin / \\
Alpaca / Crab / Eagle / Isopod / \\
Cheetah / Sea turtle / Whale / Bee / \\
Canary / Pig / Crocodile / Centipede
\end{tabular} & 
\begin{tabular}{c} 
Canoe / Helicopter / \\
Wheelchair / Ambulance / \\
Segway / Limousine 
\end{tabular} \\ \specialrule{0.7pt}{0pt}{0pt}\hline 
\end{tabular}
}
\end{table*}
\renewcommand{\arraystretch}{1}
% -------------------------------------------------
% -------------------------------------------------
\renewcommand{\arraystretch}{1.5}
\begin{table*}[h]
\centering
\caption{
Classes contained in the employed splits for CUB200 \citep{CUB200}.
}
\label{tbl:category_details_cub200}
\scalebox{0.7}{
\begin{tabular}{c||c} \specialrule{0.7pt}{0pt}{0pt}\hline

\begin{tabular}{c}
Split1\\(50)
\end{tabular} & 
\begin{tabular}{c}
Black footed Albatross / Laysan Albatross / Least Auklet / Red winged Blackbird / Yellow headed Blackbird / \\
Indigo Bunting / Spotted Catbird / Brandt Cormorant / Red faced Cormorant / Shiny Cowbird / \\
Brown Creeper / Yellow billed Cuckoo / Purple Finch / Acadian Flycatcher / Scissor tailed Flycatcher / \\
Vermilion Flycatcher / Western Grebe / Ivory Gull / Ruby throated Hummingbird / Rufous Hummingbird / \\
Green Jay / Belted Kingfisher / Pied Kingfisher / Pacific Loon / Mallard / \\
Western Meadowlark / Orchard Oriole / Scott Oriole / Whip poor Will / Loggerhead Shrike / \\
Great Grey Shrike / Brewer Sparrow / Grasshopper Sparrow / Henslow Sparrow / Le Conte Sparrow / \\
Cape Glossy Starling / Bank Swallow / Tree Swallow / Common Tern / Least Tern / \\
Philadelphia Vireo / Wilson Warbler / Pileated Woodpecker / Red bellied Woodpecker / Red cockaded Woodpecker / \\
Bewick Wren / Marsh Wren / Rock Wren / Winter Wren / Common Yellowthroat \\
\end{tabular} \\ \hline

\begin{tabular}{c}
Split2\\(50)
\end{tabular} & 
\begin{tabular}{c} 
Groove billed Ani / Crested Auklet / Parakeet Auklet / Bobolink / Lazuli Bunting / \\
Gray Catbird / Fish Crow / Gray crowned Rosy Finch / Least Flycatcher / Gadwall / \\
Blue Grosbeak / Heermann Gull / Ring billed Gull / Slaty backed Gull / Green Violetear / \\
Pomarine Jaeger / Red breasted Merganser / Mockingbird / White breasted Nuthatch / Baltimore Oriole / \\
Western Wood Pewee / American Pipit / Geococcyx / Baird Sparrow / House Sparrow / \\
Field Sparrow / Seaside Sparrow / Vesper Sparrow / White throated Sparrow / Cliff Swallow / \\
Scarlet Tanager / Summer Tanager / Elegant Tern / Forsters Tern / Green tailed Towhee / \\
Brown Thrasher / Blue headed Vireo / White eyed Vireo / Bay breasted Warbler / Black and white Warbler / \\
Golden winged Warbler / Nashville Warbler / Orange crowned Warbler / Palm Warbler / Pine Warbler / \\
Swainson Warbler / Tennessee Warbler / Bohemian Waxwing / American Three toed Woodpecker / Carolina Wren \\
\end{tabular} \\ \hline

\begin{tabular}{c}
Split3\\(50)
\end{tabular} & 
\begin{tabular}{c}
Sooty Albatross / Rhinoceros Auklet / Brewer Blackbird / Rusty Blackbird / Painted Bunting / \\
Cardinal / Chuck will Widow / Pelagic Cormorant / Bronzed Cowbird / American Crow / \\
Mangrove Cuckoo / Yellow bellied Flycatcher / Northern Fulmar / European Goldfinch / Boat tailed Grackle / \\
Horned Grebe / Evening Grosbeak / Pigeon Guillemot / Herring Gull / Western Gull / \\
Anna Hummingbird / Long tailed Jaeger / Gray Kingbird / Green Kingfisher / Horned Lark / \\
Clark Nutcracker / Brown Pelican / Sayornis / Common Raven / White necked Raven / \\
Black throated Sparrow / Chipping Sparrow / Clay colored Sparrow / Fox Sparrow / Savannah Sparrow / \\
White crowned Sparrow / Barn Swallow / Black Tern / Caspian Tern / Sage Thrasher / \\
Red eyed Vireo / Cape May Warbler / Chestnut sided Warbler / Kentucky Warbler / Mourning Warbler / \\
Prairie Warbler / Yellow Warbler / Louisiana Waterthrush / Red headed Woodpecker / Cactus Wren
\end{tabular} \\ \hline

\begin{tabular}{c}
Split4\\(50)
\end{tabular} & 
\begin{tabular}{c} 
Yellow breasted Chat / Eastern Towhee / Black billed Cuckoo / Northern Flicker / Great Crested Flycatcher / \\
Olive sided Flycatcher / Frigatebird / American Goldfinch / Eared Grebe / Pied billed Grebe / \\
Pine Grosbeak / Rose breasted Grosbeak / California Gull / Glaucous winged Gull / Blue Jay / \\
Florida Jay / Dark eyed Junco / Tropical Kingbird / Ringed Kingfisher / White breasted Kingfisher / \\
Red legged Kittiwake / Hooded Merganser / Nighthawk / Hooded Oriole / Ovenbird / \\
White Pelican / Horned Puffin / American Redstart / Harris Sparrow / Lincoln Sparrow / \\
Nelson Sharp tailed Sparrow / Song Sparrow / Tree Sparrow / Artic Tern / Black capped Vireo / \\
Warbling Vireo / Yellow throated Vireo / Black throated Blue Warbler / Blue winged Warbler / Canada Warbler / \\
Cerulean Warbler / Hooded Warbler / Magnolia Warbler / Myrtle Warbler / Prothonotary Warbler / \\
Worm eating Warbler / Northern Waterthrush / Cedar Waxwing / Downy Woodpecker / House Wren
\end{tabular} \\ \specialrule{0.7pt}{0pt}{0pt}\hline

\end{tabular}
}
\end{table*}
\renewcommand{\arraystretch}{1}
% -------------------------------------------------

% -------------------------------------------------
\renewcommand{\arraystretch}{1.4}
\begin{table*}[h]
\centering
\caption{
Classes contained in the Unknown1 and Unkonwn2 splits of MTSD \citep{MTSD}. The rest are treated as known classes.
}
\label{tbl:category_details_mtsd}
\scalebox{0.5}{
\begin{tabular}{c|c} \specialrule{0.7pt}{0pt}{0pt}\hline

\begin{tabular}{c}
Unknown1\\
(55)
\end{tabular} & 
\begin{tabular}{c}
complementary--chevron-left--g1 / complementary--chevron-right--g1 / \\
complementary--maximum-speed-limit-20--g1 / complementary--maximum-speed-limit-25--g1 / \\
complementary--maximum-speed-limit-30--g1 / complementary--maximum-speed-limit-35--g1 / \\
complementary--maximum-speed-limit-40--g1 / complementary--maximum-speed-limit-45--g1 / \\
complementary--maximum-speed-limit-50--g1 / complementary--maximum-speed-limit-55--g1 / \\
complementary--maximum-speed-limit-70--g1 / complementary--maximum-speed-limit-75--g1 / \\
information--highway-exit--g1 / information--safety-area--g2 / \\
regulatory--detour-left--g1 / regulatory--keep-right--g6 / \\
regulatory--no-overtaking--g5 / regulatory--weight-limit-with-trucks--g1 / \\
warning--accidental-area-unsure--g2 / warning--bus-stop-ahead--g3 / \\
warning--curve-left--g2 / warning--curve-right--g2 / \\
warning--domestic-animals--g3 / warning--double-curve-first-left--g2 / \\
warning--double-curve-first-right--g2 / warning--double-turn-first-right--g1 / \\
warning--falling-rocks-or-debris-right--g2 / warning--falling-rocks-or-debris-right--g4 / \\
warning--hairpin-curve-left--g1 / warning--hairpin-curve-right--g1 / \\
warning--hairpin-curve-right--g4 / warning--horizontal-alignment-left--g1 / \\
warning--horizontal-alignment-right--g1 / warning--horizontal-alignment-right--g3 / \\
warning--junction-with-a-side-road-acute-right--g1 / warning--junction-with-a-side-road-perpendicular-left--g3 / \\
warning--junction-with-a-side-road-perpendicular-right--g3 / warning--kangaloo-crossing--g1 / \\
warning--loop-270-degree--g1 / warning--narrow-bridge--g1 / \\
warning--offset-roads--g3 / warning--railroad-crossing-with-barriers--g2 / \\
warning--railroad-intersection--g4 / warning--road-widens--g1 / \\
warning--road-widens-right--g1 / warning--slippery-motorcycles--g1 / \\
warning--slippery-road-surface--g2 / warning--steep-ascent--g7 / \\
warning--trucks-crossing--g1 / warning--turn-left--g1 / \\
warning--turn-right--g1 / warning--winding-road-first-left--g1 / \\
warning--winding-road-first-right--g1 / warning--wombat-crossing--g1 / warning--y-roads--g1 / \\
\end{tabular} \\ \hline

\begin{tabular}{c}
Unknown2\\
(115)
\end{tabular} & 
\begin{tabular}{c}
complementary--both-directions--g1 / complementary--chevron-right--g3 / complementary--go-left--g1 / \\
complementary--go-right--g1 / complementary--go-right--g2 / complementary--keep-left--g1 / \\
complementary--keep-right--g1 / complementary--maximum-speed-limit-15--g1 / complementary--one-direction-left--g1 / \\
complementary--one-direction-right--g1 / complementary--turn-left--g2 / complementary--turn-right--g2 / \\
information--airport--g2 / information--bike-route--g1 / information--camp--g1 / \\
information--gas-station--g1 / information--highway-interstate-route--g2 / information--hospital--g1 / \\
information--interstate-route--g1 / information--lodging--g1 / information--parking--g3 / \\
information--parking--g6 / information--trailer-camping--g1 / regulatory--bicycles-only--g2 / \\
regulatory--bicycles-only--g3 / regulatory--do-not-block-intersection--g1 / regulatory--do-not-stop-on-tracks--g1 / \\
regulatory--dual-lanes-go-straight-on-left--g1 / regulatory--dual-lanes-go-straight-on-right--g1 / regulatory--dual-lanes-turn-left-no-u-turn--g1 / \\
regulatory--dual-lanes-turn-left-or-straight--g1 / regulatory--dual-lanes-turn-right-or-straight--g1 / regulatory--go-straight--g3 / \\
regulatory--go-straight-or-turn-left--g2 / regulatory--go-straight-or-turn-left--g3 / regulatory--go-straight-or-turn-right--g3 / \\
regulatory--keep-right--g4 / regulatory--lane-control--g1 / regulatory--left-turn-yield-on-green--g1 / \\
regulatory--maximum-speed-limit-100--g3 / regulatory--maximum-speed-limit-25--g2 / regulatory--maximum-speed-limit-30--g3 / \\
regulatory--maximum-speed-limit-35--g2 / regulatory--maximum-speed-limit-40--g3 / regulatory--maximum-speed-limit-40--g6 / \\
regulatory--maximum-speed-limit-45--g3 / regulatory--maximum-speed-limit-50--g6 / regulatory--maximum-speed-limit-55--g2 / \\
regulatory--maximum-speed-limit-65--g2 / regulatory--no-left-turn--g1 / regulatory--no-parking--g2 / \\
regulatory--no-parking-or-no-stopping--g1 / regulatory--no-parking-or-no-stopping--g2 / regulatory--no-parking-or-no-stopping--g3 / \\
regulatory--no-right-turn--g1 / regulatory--no-stopping--g2 / regulatory--no-stopping--g4 / \\
regulatory--no-straight-through--g1 / regulatory--no-turn-on-red--g1 / regulatory--no-turn-on-red--g2 / \\
regulatory--no-turn-on-red--g3 / regulatory--no-turns--g1 / regulatory--no-u-turn--g1 / \\
regulatory--one-way-left--g2 / regulatory--one-way-left--g3 / regulatory--one-way-right--g2 / \\
regulatory--one-way-right--g3 / regulatory--parking-restrictions--g2 / regulatory--pass-on-either-side--g2 / \\
regulatory--passing-lane-ahead--g1 / regulatory--reversible-lanes--g2 / regulatory--road-closed--g2 / \\
regulatory--roundabout--g2 / regulatory--stop--g1 / regulatory--stop-here-on-red-or-flashing-light--g1 / \\
regulatory--stop-here-on-red-or-flashing-light--g2 / regulatory--text-four-lines--g1 / regulatory--triple-lanes-turn-left-center-lane--g1 / \\
regulatory--truck-speed-limit-60--g1 / regulatory--turn-left--g2 / regulatory--turn-right--g3 / \\
regulatory--turning-vehicles-yield-to-pedestrians--g1 / regulatory--wrong-way--g1 / warning--added-lane-right--g1 / \\
warning--bicycles-crossing--g2 / warning--bicycles-crossing--g3 / warning--divided-highway-ends--g2 / \\
warning--double-reverse-curve-right--g1 / warning--dual-lanes-right-turn-or-go-straight--g1 / warning--emergency-vehicles--g1 / \\
warning--equestrians-crossing--g2 / warning--flaggers-in-road--g1 / warning--height-restriction--g2 / \\
warning--junction-with-a-side-road-perpendicular-left--g4 / warning--pass-left-or-right--g2 / warning--pedestrians-crossing--g4 / \\
warning--pedestrians-crossing--g9 / warning--playground--g1 / warning--playground--g3 / \\
warning--railroad-crossing--g1 / warning--railroad-intersection--g3 / warning--road-narrows-left--g2 / \\
warning--road-narrows-right--g2 / warning--roundabout--g25 / warning--school-zone--g2 / \\
warning--shared-lane-motorcycles-bicycles--g1 / warning--stop-ahead--g9 / warning--texts--g1 / \\
warning--texts--g2 / warning--texts--g3 / warning--traffic-merges-right--g1 / \\
warning--traffic-signals--g3 / warning--trail-crossing--g2 / warning--two-way-traffic--g2 / \\
\end{tabular} \\ \specialrule{0.7pt}{0pt}{0pt}\hline
\end{tabular}
}
\end{table*}
\renewcommand{\arraystretch}{1}
% -------------------------------------------------

%===========================================================
\section{Additional Experimental Results}\label{sec:more_exp_res}

\subsection{More Results of A-OSE and WI}\label{sec:more_aose_wi}
Tables~\ref{tbl:wi_aose_oi}, \ref{tbl:wi_aose_cub200}, and \ref{tbl:wi_aose_MTSD} present absolute open-set error (A-OSE) and wilderness impact (WI) for each dataset split. 
% -------------------------------------------------
\renewcommand{\arraystretch}{1.3}
\begin{table*}[t]
\footnotesize\centering
\caption{
A-OSE and WI of the compared methods in the experiment of Open Images. The same experimental setting as Table~\ref{tbl:OpenImages} is used.
}
\label{tbl:wi_aose_oi}
\scalebox{0.9}{
\begin{tabular}{c||cc|cc|cc|cc|cc}\specialrule{0.7pt}{0pt}{0pt}\hline 
& \multicolumn{10}{c}{Animal} \\ \cline{2-11}
& \multicolumn{2}{c|}{Split1} & \multicolumn{2}{c|}{Split2} & \multicolumn{2}{c|}{Split3} & \multicolumn{2}{c|}{Split4} & \multicolumn{2}{c}{mean}\\ \cline{2-11}
& A-OSE & WI & A-OSE & WI & A-OSE & WI & A-OSE & WI & A-OSE & WI \\ \hline
ORE\citep{ORE} & $23,334$ & $35.9$ & $17,835$ & $30.8$ & $22,219$ & $45.3$ & $25,682$ & $\textbf{47.0}$ & $22,268\pm 2,848$ & $39.7\pm 6.7$ \\
DS\citep{DropoutSample} & $44,377$ & $44.6$ & $28,483$ & $38.6$ & $39,592$ & $53.6$ & $42,654$ & $63.6$ & $38,776\pm 6,185$ & $50.1\pm 9.4$ \\
VOS \citep{VOS} & $\textbf{12,124}$ & $34.8$ & $21,622$ & $36.6$ & $30,988$ & $50.9$ & $23,360$ & $62.1$ & $22,024\pm 6,714$ & $46.1\pm 11.2$ \\
OpenDet\citep{Opendet} & $26,426$ & $34.9$ & $22,736$ & $27.7$ & $25,075$ & $45.6$ & $26,770$ & $56.1$ & $25,252\pm 1,585$ &  $41.1\pm 10.7$ \\
Orth.Det \citep{OrthogonalDet} & $348,918$ & $\textbf{34.1}$ & $237,609$ & $30.2$ & $331,817$ & $\textbf{36.7}$ & $322,997$ & $52.7$ & $310,335\pm 49663$ & $\textbf{38.4}\pm \textbf{9.9}$\\
FCOS \citep{FCOS} & $38,858$ & $35.5$ & $34,677$ & $37.6$ & $52,234$ & $59.4$ &$30,895$ & $49.5$ & $39,166\pm 8,053$ & $45.5\pm 9.6$ \\
Faster RCNN \citep{FasterRCNN} & $14,625$ & $30.9$ & $\textbf{11,121}$ & $\textbf{27.0}$ & $\textbf{15,745}$ & $46.8$ & $\textbf{16,260}$ & $56.7$ & $\textbf{14,438}\pm \textbf{2,314}$ & $40.4\pm 13.8$ \\
\hline\hline

& \multicolumn{10}{c}{Vehicle} \\ \cline{2-11}
& \multicolumn{2}{c|}{Split1} & \multicolumn{2}{c|}{Split2} & \multicolumn{2}{c|}{Split3} & \multicolumn{2}{c|}{Split4} & \multicolumn{2}{c}{mean}\\ \cline{2-11}
& A-OSE & WI & A-OSE & WI & A-OSE & WI & A-OSE & WI & A-OSE & WI \\ \hline
ORE \citep{ORE} & $3,143$ & $17.6$ & $3,775$ & $21.5$ & $4,483$ & $\textbf{33.7}$ & $6,654$ & $26.5$ & $4,514\pm 1,323$ & $24.9\pm 6.0$ \\
DS \citep{DropoutSample} & $4,809$ & $22.7$ & $10,617$ & $37.3$ & $16,568$ & $53.6$ & $12,107$ & $34.7$ & $11,025\pm 4,204$ & $37.1\pm 11.0$ \\
VOS \citep{VOS} & $\textbf{1,460}$ & $\textbf{12.0}$ & $\textbf{1,985}$ & $\textbf{23.9}$ & $\textbf{1,796}$ & $38.3$ & $\textbf{3,090}$ & $\textbf{20.9}$ & $\textbf{2,083}\pm \textbf{611}$ & $\textbf{23.8}\pm \textbf{9.5}$ \\
OpenDet \citep{Opendet} & $3,857$ & $19.8$ & $5,640$& $25.5$ & $10,131$ & $52.1$ & $8,893$ & $30.4$ & $7,130\pm 2,502$ & $31.9\pm 12.2$ \\
Orth.Det \citep{OrthogonalDet} & $40,685$ & $22.7$ & $67,620$ & $34.7$ & $103,278$ & $43.2$ & $86,081$ & $32.9$ & $74,416\pm 267,89$ & $33.4\pm 8.4$ \\
FCOS \citep{FCOS} & $7,700$ & $26.4$ & $10,888$ & $33.7$ & $15,395$ & $55.7$ & $22,502$ & $34.8$ & $14,121\pm 5,558$ & $37.6\pm 10.9$ \\
Faster RCNN \citep{FasterRCNN} & $3,487$ & $20.7$ & $4,291$ & $25.4$ & $6,138$ & $57.1$ & $7,760$ & $31.7$ & $5,444\pm 1,956$ & $33.7\pm 16.2$ \\
\specialrule{0.7pt}{0pt}{0pt}\hline
\end{tabular}
}
\end{table*}
\renewcommand{\arraystretch}{1}
% -------------------------------------------------

% -------------------------------------------------
\renewcommand{\arraystretch}{1.3}
\begin{table*}[t]
\footnotesize\centering
\caption{
A-OSE and WI of the compared methods in the experiment of CUB200. The same experimental setting as Table~\ref{tbl:CUB200} is used.
}
\label{tbl:wi_aose_cub200}
\scalebox{0.9}{
\begin{tabular}{c||cc|cc|cc|cc|cc}\specialrule{0.7pt}{0pt}{0pt}\hline 
& \multicolumn{2}{c|}{Split1} & \multicolumn{2}{c|}{Split2} & \multicolumn{2}{c|}{Split3} & \multicolumn{2}{c|}{Split4} & \multicolumn{2}{c}{mean}\\ \cline{2-11}
& A-OSE & WI & A-OSE & WI & A-OSE & WI & A-OSE & WI & A-OSE & WI \\ \hline
ORE\citep{ORE}         & $5,001$ & $22.6$ & $4,836$ & $22.4$ & $4,562$ & $24.1$ & $4,998$ & $19.3$ & $4,849\pm 206$ & $22.1\pm 2.0$ \\
DS\citep{DropoutSample} & $\textbf{3,231}$ & $\textbf{17.4}$ & $\textbf{3,567}$ & $\textbf{20.4}$ & $\textbf{3,356}$ & $\textbf{21.8}$ & $\textbf{3,301}$ & $\textbf{16.8}$ & $\textbf{3,363}\pm \textbf{125}$ & $\textbf{19.1}\pm \textbf{2.1}$ \\
VOS \citep{VOS}        & $4,681$ & $20.3$ & $4,535$ & $21.0$ & $4,763$ & $22.5$ & $3,681$ & $18.5$ & $4,415\pm 498$ & $20.6\pm 1.6$ \\
OpenDet\citep{Opendet} & $4,384$ & $18.6$ & $4,746$ & $21.1$ & $4,426$ & $22.6$ & $4,602$ & $18.0$ & $4,539\pm 167$ &  $20.1\pm 2.2$ \\
Orth.Det \citep{OrthogonalDet} & $170,13$ & $22.3$ & $19,942$ & $22.6$ & $19,696$ & $22.2$ & $23,888$ & $19.9$ & $20,134\pm 2,832$ & $21.8\pm 1.24$ \\
FCOS \citep{FCOS} & $15,421$ & $24.1$ & $18,334$ & $27.8$ & $21,377$ & $25.8$ &$16,822$ & $24.6$ & $17,988\pm 2,553$ & $25.6\pm 1.6$  \\
Faster RCNN \citep{FasterRCNN} & $5,898$ & $22.1$ & $6,732$ & $24.0$ & $6,289$ & $24.5$ & $6,612$ & $20.0$ & $6,382\pm 206$ & $22.7\pm 3.7$ \\
\specialrule{0.7pt}{0pt}{0pt}\hline
\end{tabular}
}
\end{table*}
\renewcommand{\arraystretch}{1}
% -------------------------------------------------
% -------------------------------------------------
\renewcommand{\arraystretch}{1.3}
\begin{table*}[t]
\footnotesize\centering
\caption{
A-OSE and WI of the compared methods in the experiments of MTSD. The same setting is used as Table~\ref{tbl:MTSD}.
}
\label{tbl:wi_aose_MTSD}
\scalebox{1.0}{
\begin{tabular}{c||cc|cc|cc|cc}\specialrule{0.7pt}{0pt}{0pt}\hline 
& \multicolumn{2}{c|}{U1} & \multicolumn{2}{c|}{U2} & \multicolumn{2}{c|}{U1+2} & \multicolumn{2}{c}{mean}\\ \cline{2-9}
& A-OSE & WI & A-OSE & WI & A-OSE & WI & A-OSE & WI \\ \hline
ORE\citep{ORE}         & $1,711$ & $5.5$ & $2,050$ & $7.0$ & $3,283$ & $11.7$ & $2,348\pm 827$ & $8.0\pm 3.3$ \\
DS\citep{DropoutSample} & $1,658$ & $6.9$ & $2,084$ & $8.4$ & $3,742$ & $15.3$ & $2,495\pm 899$ & $10.2\pm 3.7$ \\
VOS \citep{VOS}        & $1,260$ & $5.4$ & $2,003$ & $8.9$ & $3,263$ & $14.3$  & $2,175\pm 1,013$ & $9.5\pm 4.5$ \\
OpenDet\citep{Opendet} & $\textbf{722}$ & $3.8$ & $\textbf{1,146}$ & $7.8$ & $\textbf{1,868}$ & $11.6$ & $\textbf{1,245}\pm \textbf{579}$ &  $7.8\pm 3.9$ \\
Orth.Det \citep{OrthogonalDet} & $36,239$ & $\textbf{2.2}$ & $46,356$ & $\textbf{3.7}$ & $82,595$ & $\textbf{5.8}$ & $55,063\pm 24,374$ & $\textbf{3.9}\pm \textbf{1.8}$ \\ 
FCOS\citep{FCOS}       & $4,897$ & $5.7$ & $7,086$ & $7.1$ & $11,983$ & $12.8$ & $7,989\pm 3,628$ & $8.5\pm 3.8$  \\
Faster RCNN\citep{FasterRCNN} & $1,144$ & $5.5$ & $1,702$ & $7.7$ & $2,846$ & $13.2$ & $1,897\pm 868$ & $8.8\pm 4.0$ \\
\specialrule{0.7pt}{0pt}{0pt}\hline
\end{tabular}
}
\end{table*}
\renewcommand{\arraystretch}{1}
% -------------------------------------------------

\subsection{More Examples of Detection Results}\label{sec:more_vis_res}
\begin{figure*}[h]
\centering
\includegraphics[width=1.0\linewidth]{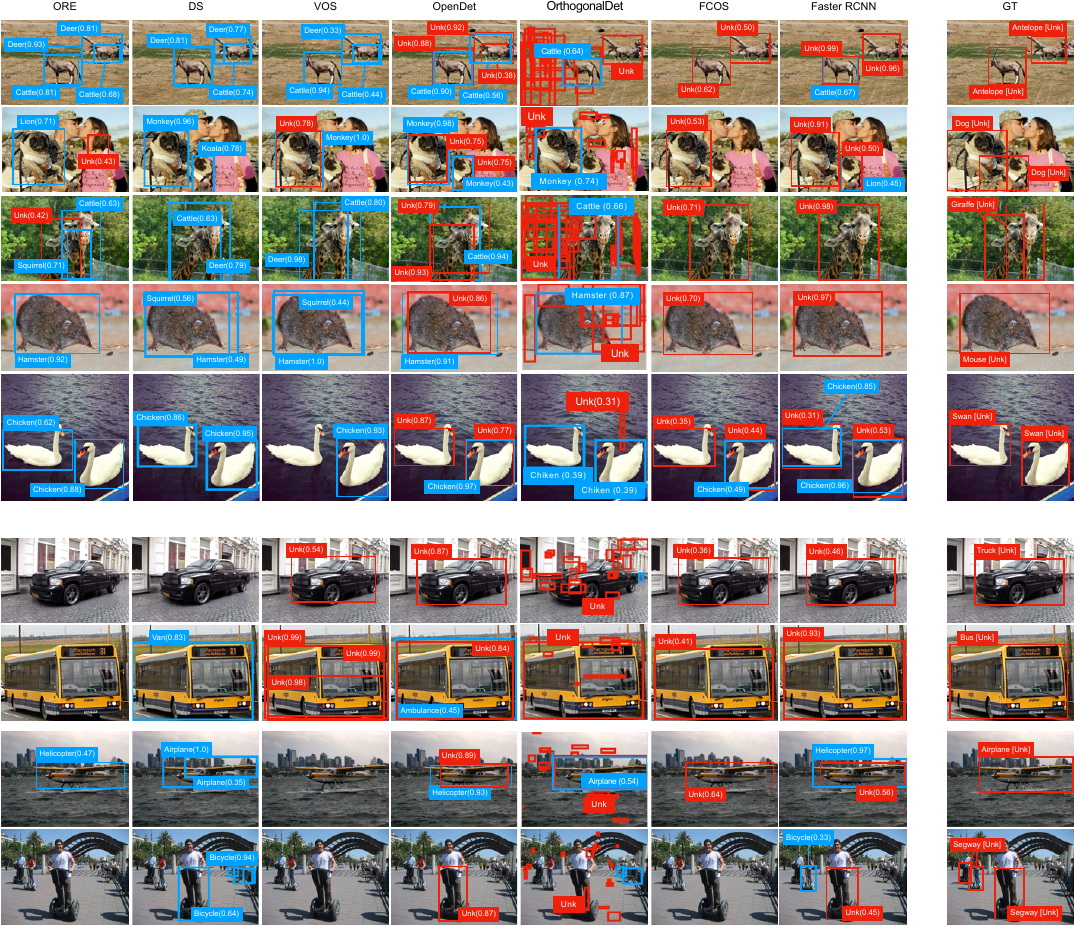}
\caption{
Examples of detection results for Open Images. Upper: the super-class is ``Animal.'' Lower: ``Vehicle.'' Red boxes represent unknown class detection, and blue boxes represent known class detection. ``Unk'' in the images stands for ``unknown''.
}
\label{fig:sup_vis_oi}
\end{figure*}

\begin{figure*}[h]
\centering
\includegraphics[width=1.0\linewidth]{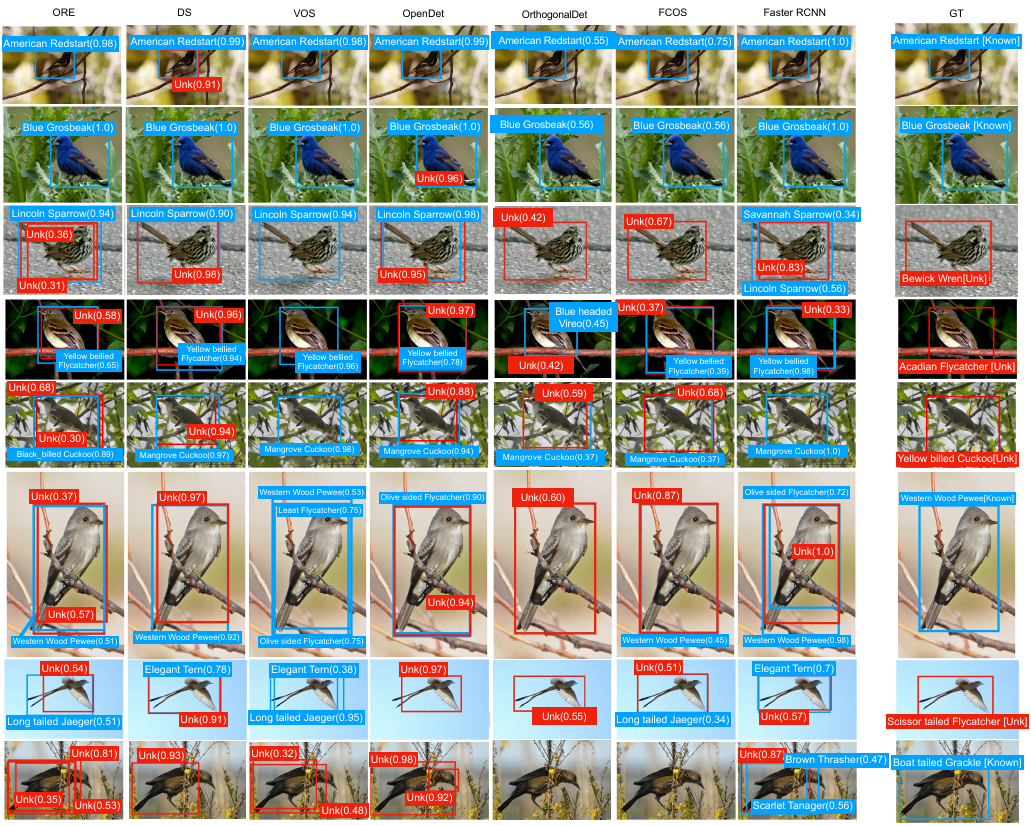}
\caption{
Examples of detection results for CUB200. See Fig~\ref{fig:sup_vis_oi} for notations. 
}
\label{fig:sup_vis_cub200}
\end{figure*}

\begin{figure*}[h]
\centering
\includegraphics[width=1.0\linewidth]{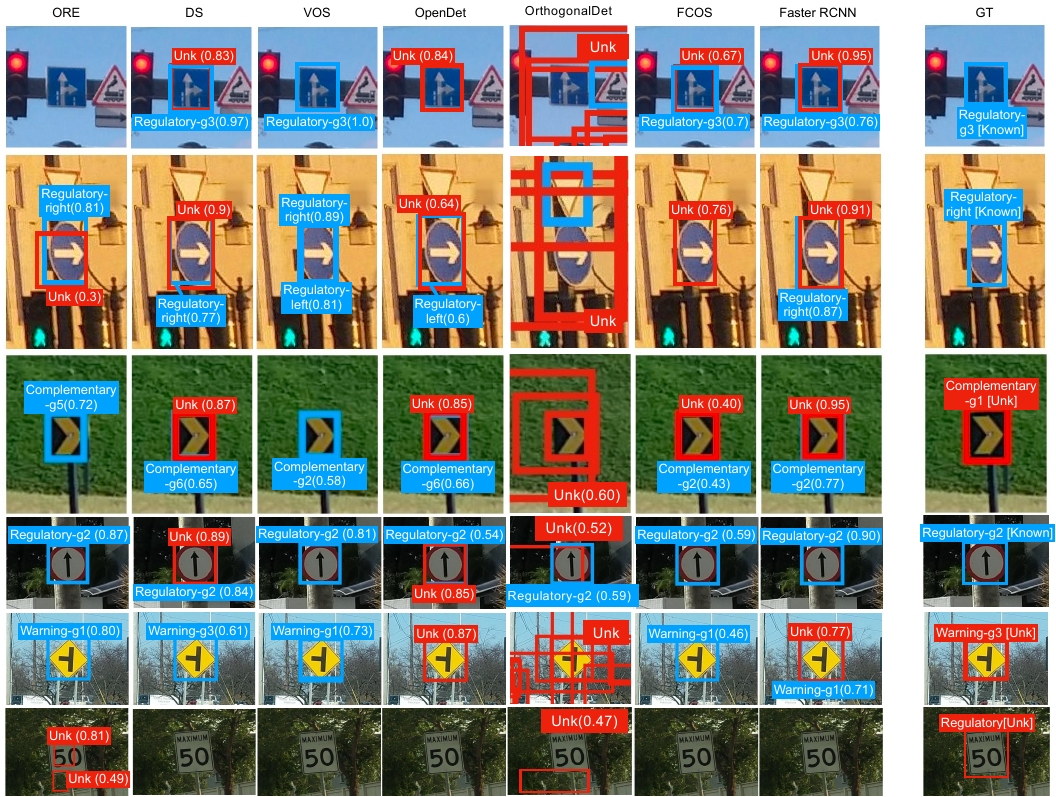}
\caption{
Examples of detection results for MTSD. See Fig~\ref{fig:sup_vis_oi} for notations.
}
\label{fig:sup_vis_mtsd}
\end{figure*}

Figures \ref{fig:sup_vis_oi}, \ref{fig:sup_vis_cub200}, and \ref{fig:sup_vis_mtsd} show more detection results for the four datasets, respectively. We only show the bounding boxes with confidence scores $>0.3$. We can observe from these results a similar tendency to the quantitative comparisons we provide in the main paper. That is, OpenDet and our baselines show comparable, limited performance in detecting unknown objects. They have the same several types of erroneous predictions, such as failures to detect unknown objects, confusion of know objects with unknown, and vice versa. 

OrthogonalDet, a state-of-the-art method for OSOD-II, exhibits a tendency to produce numerous false detections of unknown objects at random locations. This tendency contributes to the poor performance in unknown object detection, as demonstrated in Table~\ref{tbl:OpenImages}, \ref{tbl:CUB200}, and \ref{tbl:MTSD} in the main results.

As mentioned in the Sec~\ref{sec:analysis} in the main paper, they often predict two bounding boxes, significantly overlapped, with known and unknown labels for the same object instances. Their limited performance on ${\rm AP}_{unk}$, along with these failures, indicates that the existing OSOD methods will be insufficient for real-world applications.

% -------------------------------------------------
% -------------------------------------------------

% -------------------------------------------------
% -------------------------------------------------

% -------------------------------------------------
% -------------------------------------------------

\end{appendices}

\clearpage
\bibliography{main}% common bib file

\end{document}